\documentclass[11pt]{article}

\usepackage{pdflscape}
\usepackage{upquote}

\usepackage{arabtex}
\usepackage{utf8}
\makeatletter
\protected\def\begin#1{%
  \UseHook{env/#1/before}%
  \@ifundefined{#1}%
    {\def\reserved@a{%
       \@latex@error{Environment #1 undefined}\@eha}}%
    {\def\reserved@a{%
       \def\@currenvir{#1}%
       \edef\@currenvline{\on@line}%
       \@execute@begin@hook{#1}%
       \csname #1\endcsname}}%
  \@ignorefalse
  \begingroup
  \let\end\a@l@end
  \@endpefalse\reserved@a
}
\makeatother
\usepackage[final]{acl}
\usepackage{booktabs}
\usepackage{pifont}
\usepackage[x11names]{xcolor}
\usepackage{lipsum}
\usepackage{float}
\usepackage{pdfpages}
\usepackage{enumitem}
\usepackage{graphicx}
\usepackage{subcaption}
\usepackage{tabularx}
\usepackage{times}
\usepackage{latexsym}
\usepackage[T1]{fontenc}

\usepackage{microtype}
\usepackage{inconsolata}

\usepackage[utf8]{inputenc}
\usepackage{todonotes}

\usepackage{epstopdf}

\usepackage{cuted}
\usepackage{capt-of}
\usepackage{svg}
\usepackage[most]{tcolorbox}
\usepackage{seqsplit}
\usepackage{ragged2e}

\title{EDRAC: Benchmarking Arabic Dialect Reading Comprehension}

\author{
Noor Abo Mokh,\textsuperscript{1}
Kirill Chirkunov,\textsuperscript{1}
Teresa Lynn,\textsuperscript{1}
Nizar Habash,\textsuperscript{1,2}\\
\textbf{Reham Marzouk,\textsuperscript{1} 
Malik H. Altakrori,\textsuperscript{3}
Younes Samih,\textsuperscript{3}
Muhammed Abu Odeh,\textsuperscript{1}}\\
\textbf{Nour Rabih,\textsuperscript{1}
Rahaf Alshahrani,\textsuperscript{4}
Hamad Alshehhi,\textsuperscript{1}
Hamdan Al-Ali,\textsuperscript{1}}\\
\textbf{Muhra Almahri,\textsuperscript{1}
Besher Hassan,\textsuperscript{1}
Mohamed Anwar,\textsuperscript{1}
Abed Alhakim Freihat,\textsuperscript{1}} \\
\textbf{Preslav Nakov,\textsuperscript{1} 
Alham Fikri Aji\textsuperscript{1}}
\\
\textsuperscript{1}Mohamed bin Zayed University of Artificial Intelligence,\\
\textsuperscript{2}New York University Abu Dhabi,
\textsuperscript{3}IBM Research AI, \textsuperscript{4}Qassim University\\
\texttt{\{noor.abomokh,alham.fikri\}@mbzuai.ac.ae} \\}

\begin{document}

\maketitle
\begin{abstract} 
Dialectal Arabic (DA) remains under-resourced compared to Modern Standard Arabic (MSA), particularly for machine reading comprehension (MRC) and question answering (QA). Existing Arabic QA benchmarks primarily focus on formal written MSA or multiple-choice QA, with limited coverage of naturally spoken dialects. Here, we aim to bridge this gap. We introduce EDRAC, the first large-scale benchmark for dialectal Arabic machine reading comprehension (MRC) and generative QA, covering five major dialects: Egyptian, Moroccan, Emirati, Syrian, and Saudi Arabic. EDRAC contains 499 passages derived from naturally occurring spoken interactions and 4,977 corresponding QA pairs generated through a human--LLM collaborative pipeline combining iterative generation, LLM-as-a-judge evaluation, and human verification. We benchmark Arabic-centric and multilingual LLMs on EDRAC using lexical and semantic metrics. Our results reveal substantial gaps between semantic answer quality and dialectal fidelity, highlighting the limitations of existing evaluation metrics for dialectal Arabic generation. EDRAC provides a realistic and challenging MRC benchmark for future research on dialectal Arabic NLP.
\end{abstract}

\setcode{utf8}
 \vocalize
\section{Introduction}
Over the past decade, a plethora of resources have been created for Modern Standard Arabic (MSA), but there are fewer high-quality resources for Dialectal Arabic (DA),
 mainly due to the lack of written data that reflects the natural use of spoken Arabic dialects. This mainly stems from the diglossic nature of Arabic \citep{farghaly2009}, in which MSA is the primary mode of communication in written texts, while DA is used for daily interactions, resulting in fewer text-based datasets \cite{darwishHabashsurvey, dahoudialectalSurvey2025}. 

Moreover, manually creating and annotating such datasets is an expensive and laborious task. Traditionally, large DA datasets are either a translation of existing English datasets or collected from social media resources (e.g., \citet{bouamor-etal-2018-madar} and \citet{khered-etal-2025-dial2msa}). However, translations typically do not carry culturally relevant content, while social media sources do not reflect real-life conversations, spoken interactions, or natural use of the language, as speakers often alter their speech style while writing user-generated content \cite{eisenstein-2013-bad, sanguinetti-etal-2020-treebanking}. Consequently, constructing such a dataset requires transcribing audio of natural spoken conversations.

For the task of dialectal machine reading comprehension (MRC), available resources are limited to multiple-choice question answering (MCQA) tasks, (e.g., Belebele \cite{bandarkar-etal-2024-belebele} and DialectalArabicMMLU \citep{altakrori-etal-2026-dialectalarabicmmlu}). To date, Generative QA benchmarking for DA remains unexplored. Tasks such as MCQA rely on pre-defined options while EDRAC is an open-ended Question Answering task allowing for free-form descriptive, explanatory answers that  often require reasoning.

We address these gaps by presenting  EDRAC ({E}valuation of {D}ialectal Arabic {R}eading {A}nd {C}omprehension),\footnote{EDRAC \<إدراك> /{\it idraak}/ is Arabic for  `comprehension'.}$^,$\footnote{EDRAC is available under CC BY-NC-SA 4.0 license at \url{https://huggingface.co/datasets/MBZUAI/EDRAC}} a high-quality multi-dialectal MRC and Generative QA benchmark dataset for Arabic that covers Egyptian, Moroccan, Emirati, Syrian and Saudi dialects. In our curation of reading comprehension passages, we leveraged transcripts of YouTube videos, covering a variety of genres, to capture the daily interactions and 
natural flow of the dialects. We initially generated 10 QA pairs per passage; after final review, the released dataset contains 4,977 QA pairs across 499 passages. Furthermore, we  outlined the process of automatically generating these QA pairs using Large Language Models (LLMs) with a human-in-the-loop approach.
This required a two-stage framework, involving multiple steps, each including layers of filtering and reviewing, resulting in a reliable, high-quality benchmark of fully human-validated QA pairs.

Our contributions can be summarized as follows:

\begin{itemize}

\item We introduce EDRAC, the first large-scale benchmark for Dialectal Arabic Generative QA across five major dialects: Egyptian (EGY), Emirati (UAE), Moroccan (MOR), Syrian (SYR), and Saudi (KSA), built from naturally occurring spoken interactions.

\item We propose a scalable human-LLM pipeline for dialectal QA generation and validation, combining self-refinement, LLM-based evaluation, and human verification to produce high-quality, culturally grounded QA pairs.

\item We present the first comprehensive evaluation of Arabic-centric and multilingual LLMs on Dialectal Arabic Generative QA, revealing major gaps in dialectal fidelity and establishing a benchmark for future research.
 
\end{itemize}

\newcommand{\cmark}{\textcolor{black}{\ding{51}}}
\newcommand{\xmark}{\textcolor{black}{\ding{55}}}

\begin{table*}[t]
\centering
\small
\begin{tabular}{lcccc}
\toprule
\textbf{Dataset} & \textbf{MSA} & \textbf{Dialectal} & \textbf{Naturally-occurring} & \textbf{Evaluation} \\
\midrule
ACQAD \cite{sidhoum2022acqad} & \cmark & \xmark & \xmark & Extractive \\
ArabicaQA \cite{ArabicaqA_2024} & \cmark & \xmark & \xmark & Extractive\\
AraQReCC \cite{hassib-etal-2025-open}                               & \cmark & \xmark & \xmark$^\dagger$ & Conversational  \\
ARCD \& Arabic-SQuAD \cite{mozannar-etal-2019-neural}               & \cmark & \xmark & \xmark & Extractive \\
ArTrivia \cite{alrowili-vijay-shanker-2023-artrivia} & \cmark & \xmark & \xmark & Extractive \\
BALSAM \cite{almatham-etal-2025-balsam} & \cmark & \xmark & \xmark & includes QA/RC\\

MedAraBench \cite{daoud2026medarabench}                             & \cmark & \xmark & \xmark & MCQA \\
ArabicCulturalQA \cite{bhatti2026} & \cmark & \cmark & \xmark & MCQA+Generative \\
Belebele \cite{bandarkar-etal-2024-belebele}& \cmark & \cmark & \xmark & MCQA \\

DialectalArabicMMLU \cite{altakrori-etal-2026-dialectalarabicmmlu} & \cmark & \cmark & \xmark & MCQA \\
Emirati Benchmark \cite{emirati_dialect_benchmark_2026}             & \xmark & \cmark & \xmark & MCQA\\

MedQA-MA \cite{Medqa_ma}& \xmark & \cmark & \xmark & Generative \\
\midrule
\textbf{EDRAC (Ours) }                                                       & \xmark & \cmark & \cmark & Generative\\
\bottomrule
\end{tabular}
\caption{Comparison of EDRAC and other related Arabic QA benchmarks. $^\dagger$AraQReCC introduces conversational QA, but relies on translated data rather than naturally occurring speech.}
\label{tab:arabic_qa_comparison}
\end{table*}
\section{Related Work}\label{sec2:lit_review}

Existing Arabic Question Answering (QA) and Reading Comprehension (RC) benchmarks primarily focus on Modern Standard Arabic (MSA) and formal written text. For example, ARCD and Arabic-SQuAD~\citep{mozannar-etal-2019-neural} established the extractive QA setting for Arabic, while ACQAD~\cite{sidhoum2022acqad}, ArTrivia~\cite{alrowili-vijay-shanker-2023-artrivia}, ArabicaQA~\cite{ArabicaqA_2024} expanded dataset scale through Wikipedia-derived and, in some cases, automatically generated QA pairs. Other QA benchmarks include BALSAM~\citep{almatham-etal-2025-balsam} and the domain-specific MedAraBench \cite{daoud2026medarabench}. 

These benchmarks, however, remain grounded in MSA, often providing limited coverage of the variability and structural ambiguity inherent in naturally spoken Arabic defined as the language used in real-world communication.
 Recent efforts support various dialectal varieties, including the multi-dialectal Belebele \cite{bandarkar-etal-2024-belebele}, DialectalArabicMMLU \citep{altakrori-etal-2026-dialectalarabicmmlu} and ArabicCulturalQA~\citep{bhatti2026}, MedQA-MA~\citep{Medqa_ma} for Moroccan and Alyah~\citep{emirati_dialect_benchmark_2026} for Emirati. 
Most, however, rely on constructed 
written data (such as paraphrasing, translation, or controlled prompting from MSA) instead of naturally occurring dialectal speech. 
 
 EDRAC stands out by drawing on naturally occurring dialectal speech,
 capturing orthographic inconsistency, code-switching, and disfluent discourse.

\begin{figure*}[t]
    \centering
\includegraphics[width=0.9\linewidth]%
{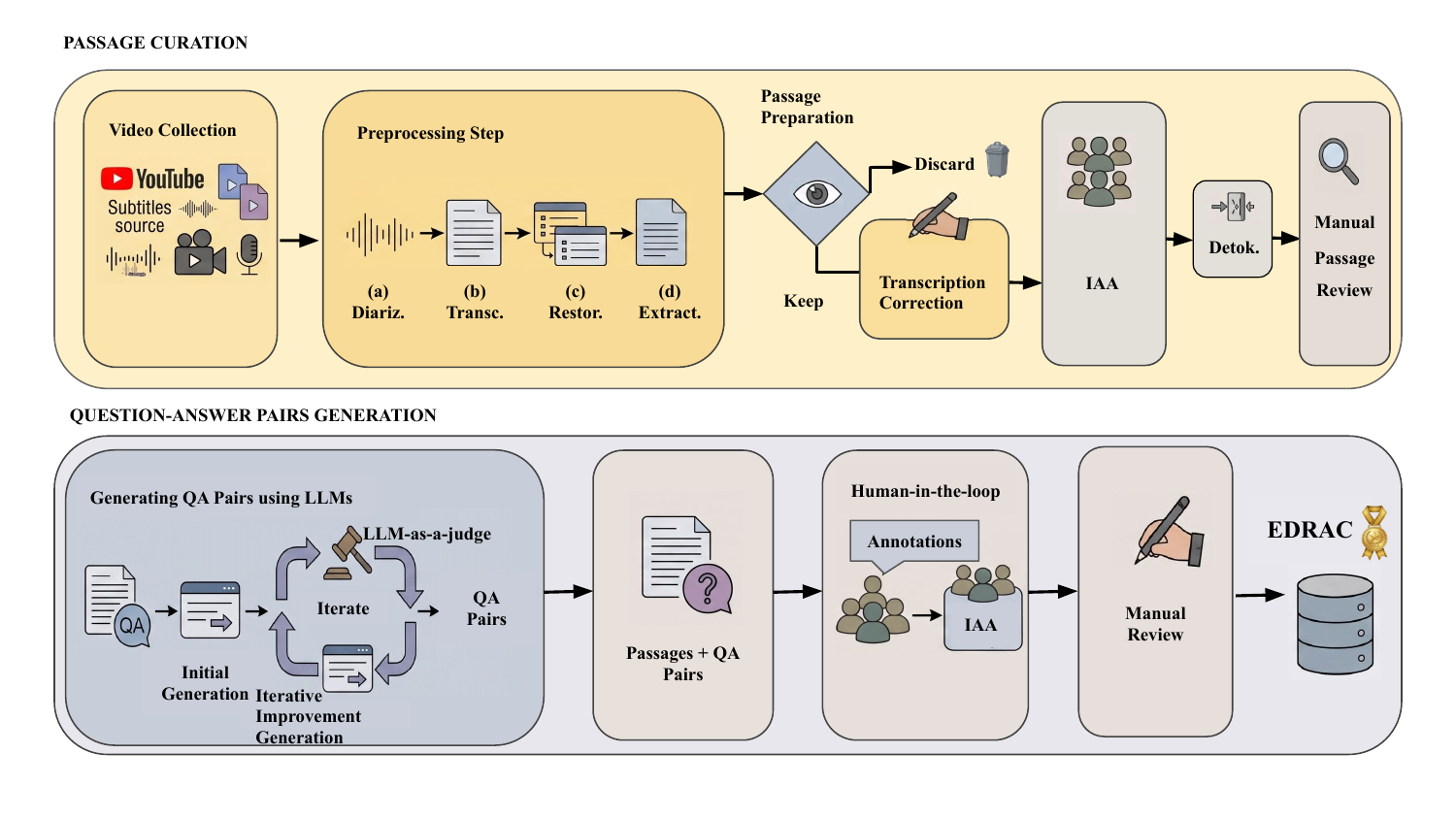}
    \caption{The end-to-end data creation pipeline of EDRAC. The top panel outlines the passage curation workflow. The bottom panel illustrates the QA generation pipeline.}
    \label{fig:pipeline}
\end{figure*}

Recent work has also explored conversational and open-domain QA. English benchmarks such as CoQA \citep{reddy-etal-2019-coqa} and QReCC \citep{anantha-etal-2021-open} introduced dialogue-aware comprehension and conversational retrieval settings, while AraQReCC  \cite{hassib-etal-2025-open} extended this paradigm to Arabic, albeit through translated conversational QA data. EDRAC departs from this paradigm by relying on authentic conversational speech rather than translated text, enabling evaluation under more realistic dialectal conditions.

To mitigate the cost of large-scale annotation, prior efforts have explored automatic QA generation  using LLMs through QAmeleon \citep{qameleon}, more specifically for multi-hop question generation (MQG) \cite{lin-etal-2024-prompting} or sequential question rewriting \cite{hwang-etal-2024-explainable}. Other QA benchmark construction efforts include languages such as French \cite{pellet-etal-2026-historiqa} for multi-hop questions on historical data and Dutch \citep{drie-etal-2026-quala} for legal-domain QA. However, automatic QA generation has not yet leveraged modern LLMs from naturally occurring Arabic speech; existing efforts such as \citet{elmadany-etal-2023-octopus} relied on encoder-decoder transformer architecture AraT5 \cite{arat5} for QA generation. 

EDRAC addresses these gaps through a robust human-in-the-loop framework in which LLMs generate candidate QA pairs 
from spoken transcripts, preserving the conversational authenticity while ensuring the data quality.
Table~\ref{tab:arabic_qa_comparison} provides a comparison of EDRAC with respect to other benchmarks.

\section{EDRAC}\label{sec3:data}
EDRAC ({E}valuation of {D}ialectal Arabic {R}eading {A}nd {C}omprehension) is a Generative QA benchmark dataset covering five major Arabic dialects: \textbf{EGY}, \textbf{UAE}, \textbf{MOR}, \textbf{SYR}, and \textbf{KSA} representing a variety of dialects across the Arab World.
We release a total of 499 passages and 4,977 corresponding QA pairs -- approximately 1000 pairs per dialect.

A central characteristic of EDRAC is its reliance on naturally occurring speech, unscripted and, in some cases, partially scripted spoken data captured in real-world settings, thereby preserving the linguistic and cultural subtleties of Arabic dialects without artificial modification. 

To construct EDRAC, we designed a hybrid pipeline combining automated processing with human verification. The process involved two main stages: \textit{passage curation} and \textit{QA pairs generation}. In the first stage (see Figure~\ref{fig:pipeline}), YouTube videos in dialectal Arabic were processed through speaker diarization, transcription alignment, text restoration, and passage extraction. This was followed by manual correction and review by native speakers of the dialects. In the second stage, the QA pairs were generated using an iterative LLM-assisted framework based on Gemini 2.5 pro, with multiple refinement steps to improve quality and reduce hallucinations. Figure~\ref{fig:pipeline} presents the complete pipeline spanning the two stages: passage curation and QA pairs generation.

The quality was ensured by an extensive manual annotation by a professional Language Service Provider: we provide annotator demographics, along with payment details in Appendix~\ref{app:demographics}. The development of comprehensive guidelines for both transcription correction and generated QAs was necessary, involving broad discussions with native speakers to ensure adequate handling of the linguistic nuances of Arabic dialects. The guidelines are available in Appendix~\ref{app:transc_guide} and  Appendix~\ref{app:qa_guide}.

\begin{figure}[t]
    \centering
    \includegraphics[trim={0cm 0cm 0cm 1cm}, clip,width=1\linewidth]{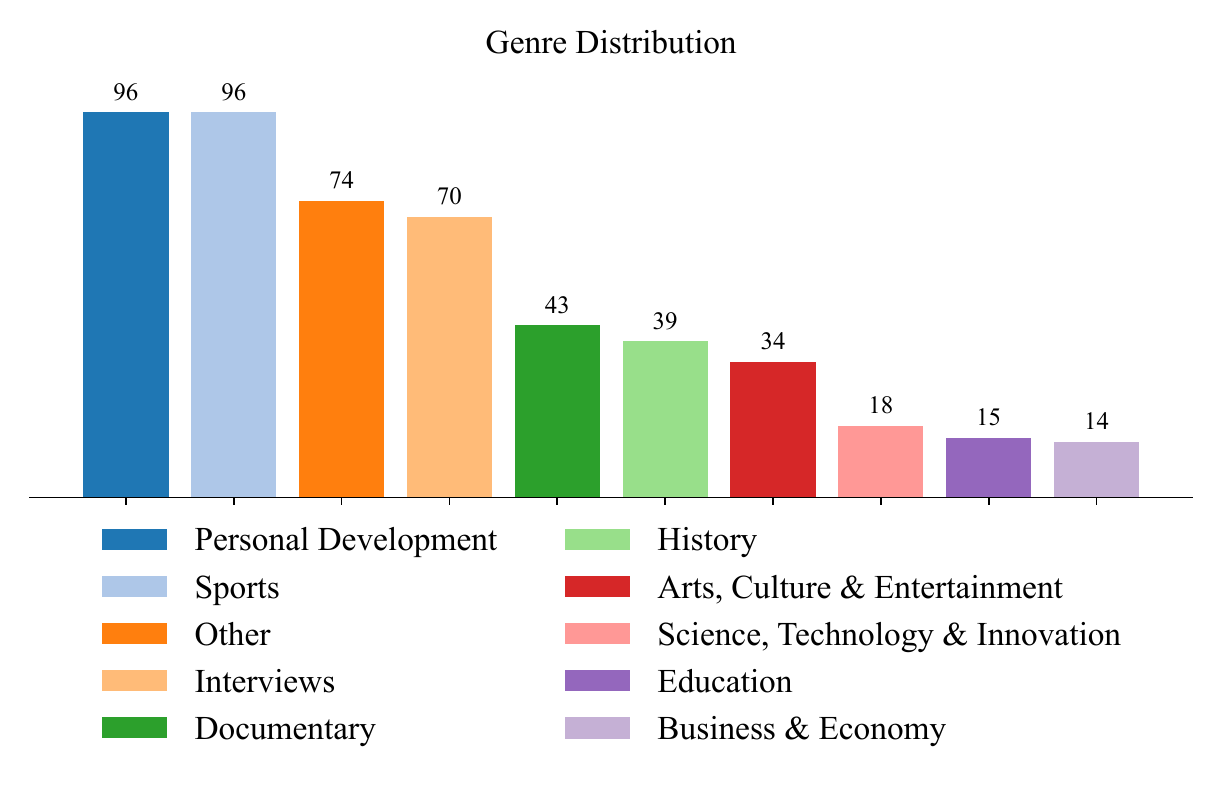}
    \caption{Genre distribution counts for the passages across dialects.}
    \label{fig:videogenres}
\end{figure}

\section{Passage Curation}
\label{sec:passagecuration}

The Arabic speech processing community has already developed extensive conversational corpora of dialectal transcribed  speech data. 
Notable datasets include ArzEn \cite{hamed-etal-2020-arzen}, STAC \cite{stac}, MASC \cite{masc} and Mixat \cite{mixat}. Nevertheless, these corpora were designed primarily for ASR or dialect identification and were thus not suitable for the purposes of our study.

Instead, we opted to curate a brand new multi-dialect transcription dataset, with post-editing focusing on the curation of reading comprehension material. We chose YouTube as our primary source for representative dialectal speech as it provided a broad variety of genres of audio content across the five dialects (see Figure~\ref{fig:videogenres}) and also allowed us to leverage pre-existing automated transcriptions. 

\paragraph{Video Collection} 
Native speakers, who were instructed to select content representative of their dialects across a variety of genres, recommended an initial list of 7 channels for their dialect and subsequent revisions and additions as necessary (see Appendix~\ref{app:channel} for the final list of channels).

Our collection was restricted to videos published between June 2024 and January 2026. Our target was 100 videos per dialect, but we sourced roughly 150 videos per dialect to allow for quality assurance filtering. We extracted the first six minutes of each video, which is equivalent to roughly 473 tokens per video passage (Table~\ref{tab:dialect_metrics}). Finally, we excluded videos containing toxic or sensitive content. We used existing YouTube subtitles as the video transcription source. Videos lacking pre-existing transcriptions were excluded.

\paragraph{Preprocessing steps} 
The preprocessing pipeline consists of four steps: \textbf{(a) Speaker diarization}: we used the NVIDIA NeMo toolkit to perform speaker diarization on the extracted audio streams. This step segments the audio by speaker identity, ensuring that multi-speaker interactions—common in media—are preserved,  \textbf{(b) Transcription}: in the transcription of speech, the tokens were timestamped to ensure alignment between the text and the original acoustic signal, \textbf{(c) Restoring text structure}: to convert fragmented ASR outputs into coherent prose, we implemented a hybrid strategy for structure restoration that involved a review stage (See Appendix \ref{app:text-restoration} for details). The automatic restoration step solely determines punctuation and segment boundaries; the original transcript tokens remain unchanged, and \textbf{(d) Text extraction}: a clean-up step discarded any passages that exceeded the passage size limit of 500 words, had missing video links, vacuous content such as promotional content and social media material such as subscription or follow requests. Finally, we extract the passages for annotation.

\paragraph{Passage Preparation} Our in-house native speakers performed a high-level review of the passages to ensure that there were no major issues. 

\paragraph{Transcription Correction} \label{sec4.1:trans_annot}
We created MS Excel-based forms for transcription correction, one passage per tab, segmented by paragraph and token-by-line split to allow close scrutiny at the token level  (see Appendix~\ref{app:transc_guide} for the annotation guidelines. We do not follow CODA \citep{habash-etal-2018-unified}).
On inspection, the quality of YouTube's Dialectal Arabic transcriptions was not sufficient for our benchmark. As such, we included a human transcription correction stage. The annotators' primary objective was to read the passage, listen to the corresponding video and correct any transcription issues. 27 annotators, representing the five dialects, collaborated in a correct-review workflow (demographics of the annotators are provided in Appendix~\ref{app:demographics}).\footnote{Annotators alternated annotation and review roles, in a streamlined correct-review process.}

 \paragraph{Detokenization and Manual Passage Review}The passages were then detokenized to restore their original structure. The final (corrected) set of transcribed passages then underwent an additional internal manual sanity check review  (e.g., removing passages or QA pairs that were non-dialectal) before finalizing the curated passage dataset.

\section{Question-Answer Pairs Generation}

Our first attempt at QA generation involved a two-week pilot study with in-house linguists. The task required extensive creative skills to ensure questions were not overly simple and could involve reasoning or deduction processes. This attempt proved so difficult and time-consuming that our next step was to explore the capability of an LLM in supporting QA generation for Arabic dialects.

\subsection{Generating QA Pairs using LLMs}

After a number of trials, \textit{Gemini-2.5-pro}~\footnote{\url{https://ai.google.dev/gemini-api/docs/models/gemini-2.5-pro}} proved to be the most reliable LLM available for automatically generating QA pairs for the dialectal passages produced in Section~\ref{sec:passagecuration}. Our initial attempts at QA pairs generation, which relied solely on a single prompt, resulted in model hallucinations (such as excessive use of MSA, and irrelevant content). To enhance the quality of the generated pairs, we introduced an iterative refinement loop. Our QA-pair generation pipeline therefore involves a three-step generate-revise framework that we outline here.

\paragraph{Step 1: Initial generation}
Given a passage, the model generates 10 QA pairs, based on a prompt devised over multiple iterative generate-review-revise rounds.\footnote{We initially explored several prompts to generate questions with varying levels of difficulty, however, the model showed inconsistency in performance in this task, as well as a reliance on verbatim answering.} The prompt instructs the model to generate reading comprehension QA pairs relevant to the provided passage, in the five dialects (see Appendix~\ref{app:initial_prompt} for the detailed prompt).

\paragraph{Step 2: LLM-as-a-judge for evaluation}
To mitigate hallucination issues from Step 1, the pipeline includes an LLM-as-a-judge to evaluate the generated pairs. This judge (also Gemini-2.5-pro) applies a fine-grained set of quality checks to the generated pairs to evaluate their quality following an approach similar to \citet{kim2024prometheus} and \citet{pombal2025mprometheus}. We instruct the model to verify if the questions are objective, unbiased, relevant, answerable and concise. We also enforce additional stylistic rules requiring questions to be in the third person and to avoid the use of linguistic or grammatical terms in the questions. We further impose additional criteria on the answers stipulating that they must be unambiguous and precise. For checks that are not met, the judge provides an explanation for the failure in addition to recommendations for improvement on the complexity of generated pairs. The full prompt is provided in Appendix~\ref{app:judge_prompt}. QA pairs that fail Step 2 of the generation framework enter into an iterative improvement loop (Step 3).

\paragraph{Step 3: Iterative improvement generation}\label{improv_prompt}
 The same model is again provided with: the passage, the failed pairs from Step 2, and more crucially, the judge's feedback on each pair’s quality. The model is explicitly instructed to correct any issues flagged during Step 2, while preserving the dialect characteristics of the passage. Steps 2 and 3 are repeated for a maximum of five times, or until the evaluated pair passes all quality checks. See Appendix~\ref{app:improv_prompt} for the full prompt.

\begin{figure*}[t]
    \centering
\includegraphics[width=1\linewidth]{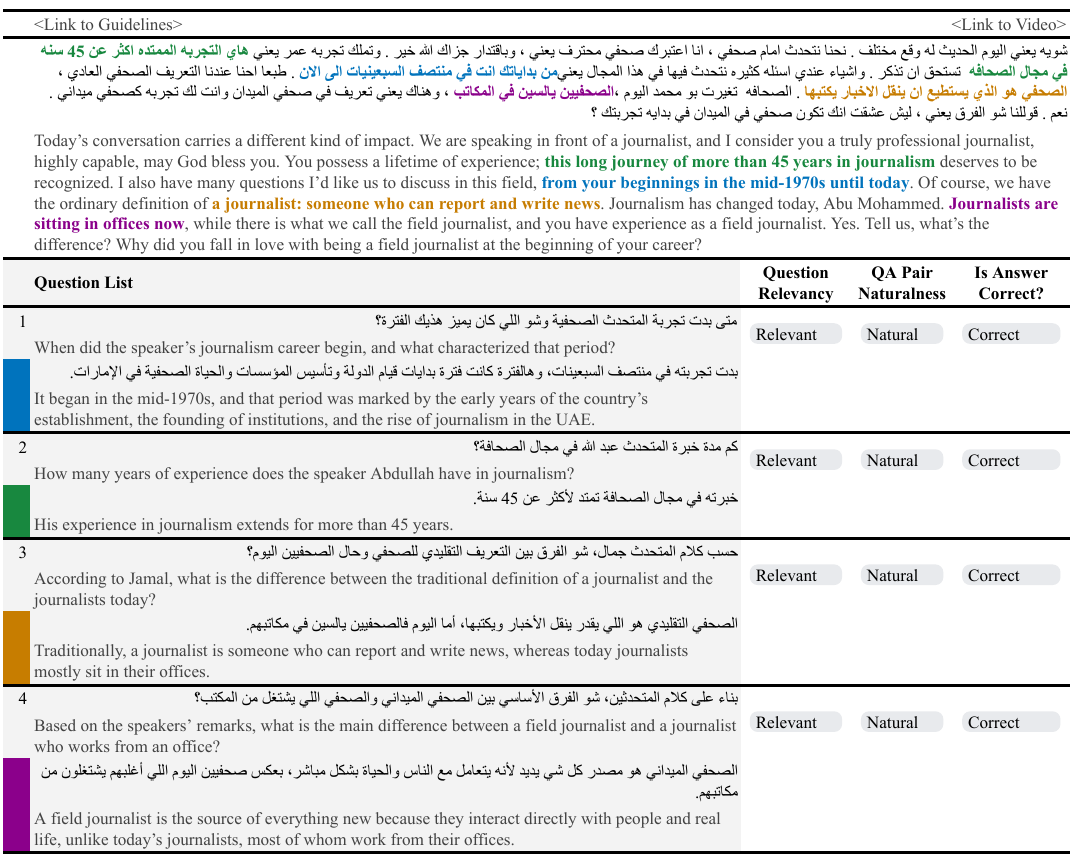}
    \caption{Example of the EDRAC Google Sheets annotation interface, showing a shortened snippet of the Arabic transcript segment, its associated question–answer pairs, annotation labels, and color-coded evidence spans to guide annotators. English translations are included only for reader clarity and are not shown to annotators.}
    \label{fig:qa_annotation_example}
\end{figure*}

\begin{table}[htbp]
\centering
\small
\setlength{\tabcolsep}{2pt}
\begin{tabular}{lrrrrrrr}
\toprule
\textbf{Dialect} & \textbf{Pass.} & \textbf{Toks} & \textbf{Sents.} & \textbf{Par.} & \textbf{Tok/} & \textbf{Sent/} & \textbf{Par./} \\
& & &  &  & \textbf{Pass.} & \textbf{Pass.} & \textbf{Pass} \\
\midrule
EGY     & 100 & 46,655 & 1,989 & 603 & 466.6 & 19.9 & 6.0 \\
MOR     & 100 & 50,174 & 2,643 & 755 & 501.7 & 26.4 & 7.6 \\
KSA     & 100 & 47,632 & 2,109 & 685 & 476.3 & 21.1 & 6.9 \\
SYR     &  99 & 42,821 & 3,057 & 1,217 & 432.5 & 30.9 & 12.3 \\
UAE     & 100 & 48,831 & 3,047 & 1,268 & 488.3 & 30.5 & 12.7 \\
\midrule
\textbf{Overall} & \textbf{499} & \textbf{236,113} & \textbf{12,845} & \textbf{4,528} & \textbf{473.17} & \textbf{25.74} & \textbf{9.07} \\
\bottomrule
\end{tabular}%
\caption{Post-review textual statistics and metrics across all dialects for passages, tokens, sentences and paragraphs.}
\label{tab:dialect_metrics}
\end{table}

\subsection{Human-in-the-Loop} \label{annot_guidelines_desc}

As benchmark data cannot be created through automated approaches alone, it is crucial to include a \textit{human-in-the-loop}. This validation step enables assessment of Gemini 2.5 pro's capability in generating  QA pairs and also flagging of LLM-related issues~\cite{Huang-hallucinations}, such as hallucinations and inaccuracies.  

\paragraph{Dialectal Annotation Guidelines} Manual evaluation of the generated QA pairs was based on defined criteria outlined in our annotation guidelines (see Appendix~\ref{app:qa_guide}). These guidelines were refined and finalized through an iterative process of pilot testing and revising. 
The first draft of the guidelines was tested in a pilot on Egyptian data, but differed significantly in terms of label categories, as it involved an attempt to generate QAs with various levels of difficulty - a categorization that proved difficult to define in terms of  prompt instructions.

In the second pilot, three native Emirati speakers reviewed 10 automatically generated QA pairs across five passages. 

The results and follow-up session revealed noticeable disagreement among annotators driven primarily by annotators' regional linguistic variation and daily linguistic habits. Following the pilot, the guidelines were updated to improve the definition and categories of `naturalness'. We also added clearer explanation on what was acceptable as dialectal variations and how to handle code-switching. At this stage, five different sets of guidelines  were deemed more appropriate, one for each dialect. Each version differs only in the dialect-specific examples used.

\paragraph{Annotation Set-up}Dialect teams of two were assigned 100 passages and their 10 corresponding QA pairs (1000 QA pairs per dialect team). Each annotator annotated 60 passages with an overlap of 20 to allow for an IAA study. In the annotation file, each generated QA pair is cross-referenced with automatic color-coding highlighted spans within the passage to support efficiency in finding the relevant section(s) of the passage.  We also include a link to the video relevant to each passage. See Figure~\ref{fig:qa_annotation_example} for a visual example. The ultimate goal was to assess the LLM's output in terms of relevancy, naturalness, and correctness, with an ``ideal'' result of \textit{Relevant}, \textit{Natural} and \textit{Correct} labels.

\begin{enumerate}
    \item \textbf{Relevancy}: Is the question relevant to the content of the passage? (Relevant, Irrelevant)
    \item \textbf{Naturalness}: How well do the generated QA pairs represent the dialect as it is naturally spoken by native speakers? (Natural, Somewhat Awkward, Unnatural, Non-dialectal.)
    \item \textbf{Correctness}: Does the answer meet the requirements of being  factually accurate and supported directly by the passage, or concluded or inferred from information in the passage? (Correct, Incorrect, Invalid).
\end{enumerate}

\paragraph{IAA} As an additional measure of quality, we conduct inter-annotator agreement  using Cohen’s Kappa coefficient~\cite{Cohen1960ACO} on 20 passages per dialect. 
While the annotations achieved high observed agreement across the categories, the resulting Kappa scores remained comparatively low due to the highly skewed label distribution within our dataset (also known as the ``Kappa Paradox''  \cite{feinstein_high_1990}).  We therefore calculated PABAK-OS scores (a measure used for linguistic analysis in \citet{anzovino2018automatic}). The results showed high agreement across dialects. See Appendix~\ref{sec:IAA_QA} for all results. 

\begin{table}[htbp]
\centering
\small
\begin{tabular}{lrrr}
\toprule
\textbf{Dialect} & \textbf{Relevant} & \textbf{Natural} & \textbf{Correct} \\
\midrule
EGY & 100\% & 96\% & 98\% \\
MOR & 98\% & 99\% & 96\% \\
KSA & 99\% & 95\% & 96\% \\
SYR & 97\% & 98\% & 90\% \\
UAE & 100\% & 98\%  & 98\% \\
\bottomrule
\end{tabular}%

\caption{Pre-review label distribution summary. The three dimensions were evaluated independently.}
\label{tab:label_distribution}
\end{table}

\paragraph{Manual Review} 

We performed a final round of manual reviews by in-house native speakers. Spot checks were carried out on the quality of annotations for each dialect. 209 QA pairs out of 5,000 (4.2\%) were flagged. We classify the issues flagged into: (a) Fix (b) Discard (c) No change  (see 
Table~\ref{tab:dialect_fixes_total} in Appendix~\ref{app:final_stats} for the statistics of the data, and Appendix~\ref{app:flagged_issue} for an example of a flagged issue).

Therefore the Total Flagged is a very small percentage of the entire dialectal set of 5000 QA pairs. At this stage, 23 QA pairs were discarded, including one full passage entry, resulting in 499 passages. It is worth noting that a significantly low number of LLM-generated issues were flagged by our annotators. These issues generally related to incorrect answers, misinterpretation of the question, gender confusion (this affecting morphology of several tokens), partial correctness (e.g., only retrieving part of a list) and hallucinated answers (not found in the passage). The distribution of the ``ideal'' response labels (\textit{Relevant}, \textit{Natural} and \textit{Correct}) for the pre-review generated pairs are shown in Table~\ref{tab:label_distribution}. Table~\ref{tab:err-categories} provides a summary of the types of errors encountered during the manual review.

\begin{table}[t]
\centering
\small
\begin{tabular}{lr}
\toprule
\textbf{Error Category} & \textbf{QA pairs \%} \\
\midrule
Incorrect Annotation & 1\% \\
Language naturalness & <1\% \\
Model answer errors & <1\% \\
Question quality issues & <0.1\% \\
Cultural understanding errors & <0.1\% \\
Transcription quality issues  & <0.1\% \\
 &  \\
\bottomrule
\end{tabular}
\caption{Pre-review percentage of QA error categories.}
\label{tab:err-categories}
\end{table}

The results of the manual review demonstrate two insights. First, Gemini 2.5 Pro can be reliably used in an iterative-improvement loop approach to generate these 5 dialectal varieties of Arabic. Second, we confirm the quality of the dataset.

\begin{figure*}[t!]
    \centering
    \noindent
    \includegraphics[width=\linewidth]{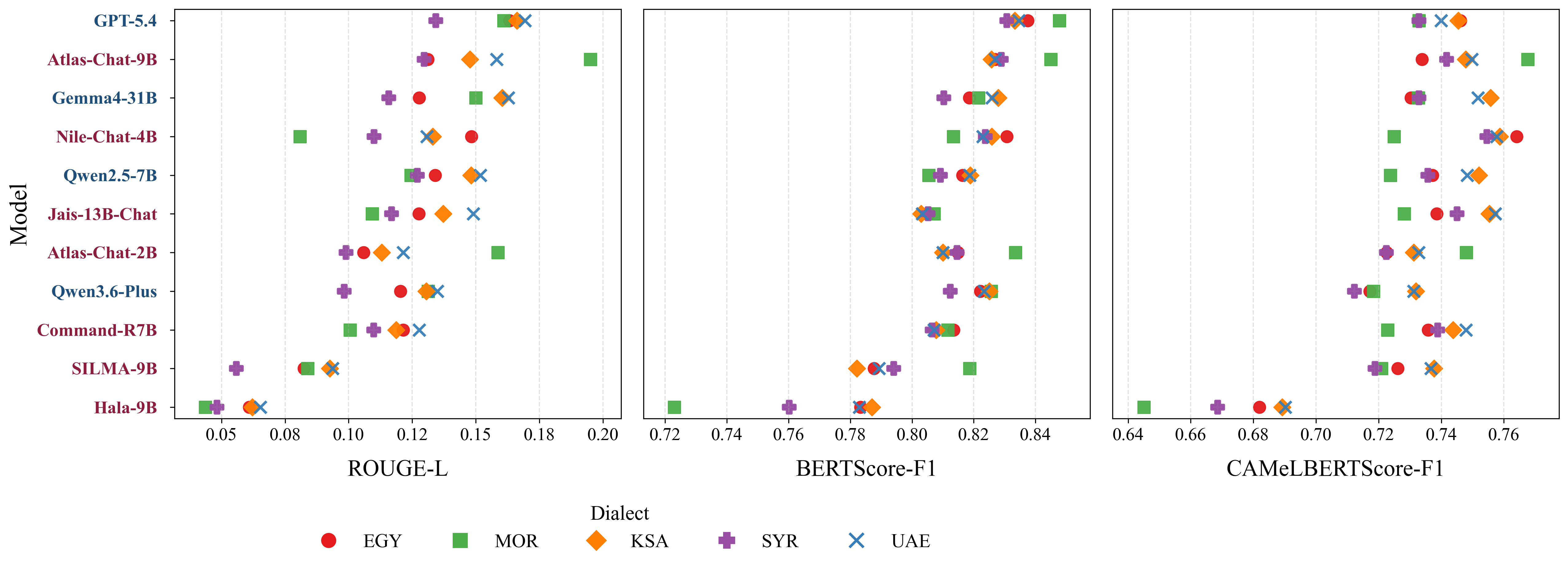}
    \caption{Model performance across five Arabic dialects under three different evaluation metrics. \textcolor{SteelBlue4}{Blue} 
    model names indicate general multilingual models, whereas \textcolor{Firebrick4}{Red} model names indicate Arabic-focused models.}
    \label{fig:overall}
    
\end{figure*}

\subsection{Data Challenges}

Developing a high-quality dataset for DA comes with challenges, stemming from the language's diglossic nature, the vast number of varieties, regional variation that is not well-documented, and attitudes towards these varieties. 

One issue that influenced both the design of the guidelines and the annotation process is the orthographic variation among speakers of Arabic as there is no standardized orthography that is commonly adopted among speakers. Speakers' spelling choices are influenced by the phonological representation of the word or MSA orthographic rules. In our guidelines, we emphasized that spelling variants are acceptable when there is no meaning change.

The definition of naturalness posed another challenge. The Emirati pilot (see Section~\ref{annot_guidelines_desc}) revealed a high disagreement among speakers. An underlying reason for judgment variations was the annotators' own regional dialectal preferences, and assumptions about their dialect. This was noticeable in the `naturalness' category. Additionally, each Arabic dialect has its own sub-varieties that can vary lexically and phonologically. Not all speakers are familiar with such sub-varieties, and may therefore be 
biased against them.
From an annotation point of view, this means the annotators' judgments can be biased and may not reflect the true nature of the dialects. The subjective judgments and regional attitudes directly impacted label distribution in the Emirati pilot. Annotators were often hesitant to label unfamiliar sub-varieties as `natural'. To counter this, our guidelines help annotators to look past their sub-variety bias. We have also emphasized that variation is a natural linguistic phenomenon across speakers.

Additionally, some of the passages contained high-register language characteristic of formal media contexts (e.g., news).
Prestige bias led some annotators to regard such language as unnatural.

We noted in our guidelines that these forms  reflect the linguistic nuances of the native dialect and should be labeled ‘natural’.

On the modeling side, we observed dialect mixing in an earlier attempt at generating QA pairs e.g., Syrian with Palestinian, Emirati with Saudi. This is not unusual, as models are often biased. We address this in our prompt by specifying both the dialect and the country where it is spoken.

\section{Experimental Setup}\label{sec5:exps}

We use the EDRAC dataset introduced in Section~\ref{sec3:data} to establish a benchmark and evaluate a set of Arabic-centric and multilingual LLMs on their ability to answer dialectal RC questions. In choosing these models, we focused on dialectal models Atlas-Chat-2B, Atlas-Chat-9B, Nile-Chat-4B, Hala-9B and Jais-13B-Chat that support Moroccan, Egyptian, Saudi and Emirati dialects, respectively. In addition, we evaluate a number of multilingual models that are both open-weight models (Qwen2.5-7B, Gemma4-31B and SILMA-9B) and closed-weight models (GPT-5.4 and Qwen3.6-Plus). Details are provided in Table~\ref{tab:List_models} in Appendix~\ref{app:experimental}. 

For the QA setup,\footnote{Source code is available at \url{https://github.com/mbzuai-nlp/EDRAC}.} we used OpenRouter~\cite{openrouter2026}, an online API that provides a unified access to different models using one standard interface. The models were prompted with:

\begin{small}
\noindent\texttt{<<Given the following passage, answer the accompanying question. \\Passage: \{passage\}\\
Question: \{question\} \\
Answer:>>}\\
\end{small}
We provide the prompt to the models in English rather than Arabic. This aligns with the standard practice of using English as the language of the prompt, especially when benchmarking a highly diverse suite of both multilingual and Arabic-centric LLMs. For example, \citet{Aljagthami2025EvaluatingLL} and \citet{dey2024betteraskenglishevaluation} have demonstrated that models show a more consistent performance when prompted in English.

\subsection{Evaluation Metrics}

We report model performance on answering questions using three metrics: (1) ROUGE - a token-based metric (whitespace tokenization), (2) BERTScore model (mDeBERTa-based implementation), and (3) CAMeLBERTScore which is an Arabic-specific implementation. ROUGE~\citep{lin-2004-rouge} is typically used for the evaluation of text summarization and machine translation tasks \citep{10.1145/3641289}. It is a token-based metric and can only capture exact lexical matches between the gold and generated texts.

We also report the performance using BERTScore-F1, which is better suited to measuring the semantic similarity. 
We evaluate with BERTScore~\cite{zhang2020bertscore} using two different encoder-only language models: mDeBERTa, a multilingual variant of DeBERTaV3~\cite{he2023debertav3improvingdebertausing}, and CAMeLBERT~\cite{inoue-etal-2021-interplay}, a model fine-tuned for Arabic dialects. We use CAMeL-Lab/bert-base-arabic-camelbert-mix. 

We provide results in Figure~\ref{fig:overall}. The model order in the plot is based on average performance across dialects, under three metrics, sorted in descending order (1st row is the best performing model). 

\section{Results and Analysis}

Figure~\ref{fig:average_perf_per_dial} illustrates model aggregate performance on the EDRAC dataset across various dialects, evaluated using three distinct metrics (also see Appendix~\ref{app:results} for the results). According to ROUGE-L and BERTScore-F1, the flagship GPT-5.4 achieved the highest average performance across all dialects. While the model's proprietary nature limits our analysis, we can only speculate that this advantage stems from its massive parameter size and extensive training data.

Among the Arabic-focused models, Atlas-Chat-9B achieved solid overall performance. Notably, the Atlas models exhibited exceptionally strong performance on the Moroccan dialect, which aligns with their Moroccan-focused training data. Furthermore, Nile-Chat-4B demonstrated remarkable efficiency, outperforming significantly larger Arabic-centric models such as Jais-13B-Chat, SILMA-9B, and Hala-9B despite its compact 4B parameter size.

While Gemma4-31B also delivered strong results, its massive scale makes it less practical for deployment, especially considering it was outperformed by the much smaller Atlas-Chat-9B in most scenarios. Conversely, Qwen3.6-Plus exhibited a degradation in performance compared to its predecessor, suggesting a weaker Arabic performance in this evaluation setting.

Lastly, we observed that standard metrics like ROUGE-L and BERTScore over-reward models for broad semantic similarity while failing to capture dialectal nuances, creating a misleading leaderboard that does not always accurately reflect true dialectal proficiency. For instance, when evaluated using the Arabic-specific CAMeLBERTScore-F1 metric, GPT-5.4 dropped from 1st to joint 5th with Qwen2.5-7B. This suggests that while the model generates semantically accurate responses, it struggles with dialectal constructions and specific lexical overlap.

\subsection{Human Evaluations}

\begin{table}[t!]
\centering
\small
\tabcolsep4pt
\begin{tabular}{lrrrr}
\toprule
\textbf{Dialect} & \textbf{Rouge} & \textbf{BertScore} & \textbf{Accuracy} & \textbf{Naturalness} \\
\midrule
\textbf{EGY} & 13\% & 82\% & 85\% & 32\% \\
\textbf{MOR} & 11\% & 85\% & 95\% & 50\% \\
\textbf{KSA} & 13\% & 81\% & 95\% & 25\% \\
\textbf{SYR} & 8\% & 79\% & 82\% & 42\% \\
\textbf{UAE} & 13\% & 81\% & 100\% & 5\% \\
\bottomrule
\end{tabular}
\caption{Comparison of Automatic and Human Evaluation Scores Across Five Arabic Dialects. Human evaluation scores (Accuracy  and Naturalness) are based on a 40-example sample per dialect drawn equally from GPT-5.4 and Hala-9B outputs.}
\label{tab:human-eval-gpt-hala}
\end{table}

We conducted human evaluations on two models to further investigate our findings. We selected 20 examples from GPT-5.4 output (best performing model) and 20 examples from Hala-9B output (worst performing model) for each of the dialects. To ensure a representative sample across genres, we selected two questions from 10 different passages within each dialectal set. We evaluate the model outputs on two dimensions: accuracy and naturalness (see Appendix \ref{app:hum_eval_guide} for annotation guidelines). 
The accuracy measures semantic equivalence with a reference answer. To avoid bias and lexical priming during evaluations, we do not provide the passages to the annotators.

Table~\ref{tab:human-eval-gpt-hala} shows a  clear discrepancy between automatic and human metrics. Standard metrics like ROUGE penalize lengthy or elaborate answers even when they are representative of accurate human responses. Conversely, while semantic-similarity metrics like BERTScore show high scores across the board, they fail to capture critical issues with the "dialectness" of the generated text. For example, for some correct answers, the annotators also labeled the answer as non-dialectal.

\section{Conclusion and Future Work}

We introduced EDRAC, the first large-scale benchmark for dialectal Arabic machine reading comprehension and generative question answering across five Arabic dialects. EDRAC is grounded in naturally occurring spoken interactions and constructed through a scalable human--LLM collaborative pipeline of iterative generation, LLM-as-a-judge evaluation, and human verification. Our experiments with Arabic-centric and multilingual LLMs reveal substantial gaps between semantic answer quality and dialectal fidelity, highlighting limitations of current evaluation metrics for dialectal Arabic. Our roadmap for building comparable resources can help other low-resource spoken varieties to address the challenge of data scarcity.

In future work, we plan to expand EDRAC to additional dialects and more challenging reasoning settings, including conversational and multi-hop QA. We also aim to investigate dialect-sensitive evaluation metrics and extend our framework to other low-resource spoken languages.

\section*{Limitations}

EDRAC covers five major Arabic dialects, but it does not capture the full linguistic diversity of the Arabic-speaking world, including finer-grained regional and sociolectal variation. In addition, although the dataset is grounded in naturally spoken interactions, some source content may still contain partial scripting or speaker self-monitoring typical of online media. 

Our QA generation pipeline relies on LLM-assisted generation and evaluation, which may introduce residual biases or stylistic artifacts despite multiple rounds of human verification and quality control. We also note that there are many high-quality Arabic-centric models available that support Arabic, yet we limited our choice of models to a representative and diverse sample for the purposes of our study. 

Finally, while we evaluate a diverse set of Arabic-centric and multilingual LLMs, the rapidly evolving landscape of language models means that future systems may exhibit different performance trends on the benchmark.

\section*{Ethics and Broader Impact}

EDRAC was constructed from publicly available YouTube content using only limited transcript excerpts necessary for the research objectives. The dataset is released under the CC BY-NC-SA 4.0 license for non-commercial use and is intended to be solely an evaluation benchmark. To preserve benchmark integrity, it should not be included in pretraining or fine-tuning data.

To reduce potential harms, we excluded videos containing sensitive, toxic, or explicit content during data collection and applied multiple layers of human review throughout the annotation pipeline. The dataset is designed to support research on dialectal Arabic NLP, particularly for underrepresented spoken varieties that are often overlooked in current language technologies.

At the same time, dialectal data may contain regional, social, or cultural biases that reflect naturally occurring speech. While we employed native-speaker annotators and quality-control procedures to improve consistency and linguistic authenticity, residual biases and annotation subjectivity may still remain. In addition, the use of LLM-assisted QA generation may introduce stylistic artifacts or model-specific biases despite human verification.

We hope EDRAC contributes to more inclusive and representative Arabic NLP systems by encouraging research on dialect-aware language technologies and more robust evaluation methods for spoken language understanding. We used AI writing assistance within the scope of ``Assistance
purely with the language of the paper'' described in the ACL Policy on Publication Ethics.

\section*{Acknowledgments}
This work was conducted as part of the IBM–MBZUAI AI Center of Excellence. 
The research is in part with support from Google.org and the Google Cloud Research Credits program for the Gemini Academic Program through a grant to New York University Abu Dhabi's CAMeL Lab.
The authors gratefully acknowledge the contributions of Amr Keleg and Samar Magdy for their assistance during the data curation process.

\bibliography{custom}

\appendix

\begin{onecolumn}
\newpage
\section{Results}\label{app:results}

\begin{table*}[hpb]
\centering
\vspace{0.2cm}
\setlength{\tabcolsep}{3.5pt}
\resizebox{\textwidth}{!}{%
\small
\begin{tabular}{p{3cm}l *{15}{c}}
\toprule
 & \multicolumn{3}{c}{\textbf{EGY}} & \multicolumn{3}{c}{\textbf{MOR}} & \multicolumn{3}{c}{\textbf{KSA}} & \multicolumn{3}{c}{\textbf{SYR}} & \multicolumn{3}{c}{\textbf{UAE}} \\
\cmidrule(lr){2-4} \cmidrule(lr){5-7} \cmidrule(lr){8-10} \cmidrule(lr){11-13} \cmidrule(lr){14-16}
\textbf{Model} & \textbf{RL} & \textbf{BF$^D_1$} & \textbf{BF$^C_1$} & \textbf{RL} & \textbf{BF$^D_1$} & \textbf{BF$^C_1$} & \textbf{RL} & \textbf{BF$^D_1$} & \textbf{BF$^C_1$} & \textbf{RL} & \textbf{BF$^D_1$} & \textbf{BF$^C_1$} & \textbf{RL} & \textbf{BF$^D_1$} & \textbf{BF$^C_1$}\\
\midrule
GPT-5.4 & 16.3 & 83.8 & 74.6 & 16.1 & 84.8 & 73.3 & 16.6 & 83.3 &  74.5& 13.4 & 83.1 & 73.3 & 16.9 &83.5  &74.0  \\
Qwen3.6-Plus & 12.0 & 82.2 & 71.7 & 13.1 & 82.6 & 71.8& 13.1 & 82.5 & 73.2 & 9.82 & 81.2 & 71.2 & 13.5 & 82.3 & 73.1 \\
Qwen2.5-7B & 13.4 & 81.6 & 73.7 & 12.5 & 80.5 & 72.4 & 14.8 & 81.9 & 75.2 & 12.7 & 80.9 & 73.6 & 15.2 & 81.9 & 74.8 \\
Gemma4-31B & 12.8 & 81.9 & 73.0 & 15.0 & 82.2 & 73.3 & 16.0 & 82.8 & 75.6 & 11.6 & 81.0 & 73.3 & 16.3 & 82.6 & 75.2 \\
Jais-13B-Chat & 12.8 & 80.5 & 73.9 & 10.9 & 80.7 & 72.8 & 13.7 & 80.3 & 75.5 &  11.7&  80.5& 74.5 &14.9  &80.3  & 75.7 \\
SILMA-9B & 8.24 & 78.8 & 72.6 & 8.39 & 81.9 & 72.1 & 9.26 &78.2  & 73.8 & 5.58 & 79.4 & 71.9 & 9.36 & 78.9 &73.7  \\
Command-R7B & 12.2 & 81.4 & 73.6 & 10.1 & 81.2 & 72.3 & 11.9 & 80.8 & 74.4 & 11.0 & 80.7 &73.9  &12.8  &80.7  & 74.8 \\
Nile-Chat-4B& 14.8 & 83.1 &76.4  & 8.09 & 81.3 & 72.5 & 13.3 & 82.6 & 75.9 & 11.0 & 82.4 & 75.5 & 13.1 &82.3  & 75.8 \\
Atlas-Chat-2B & 10.6 &81.5  &72.3  &15.9  &83.4  &74.8  & 11.3 & 81.0 & 73.1 & 9.90 &81.5  &72.3  &12.1  &81.0  &73.3  \\
Atlas-Chat-9B&  13.1& 82.7 &73.4  &19.5  &84.5  &76.8  &14.8  &82.6  &74.8  & 13.0 &82.9  &74.2  &15.9  &82.7  &75.0  \\
Hala-9B & 6.09 & 78.3 & 68.2 & 4.37 & 72.3 & 64.5 &6.21  &78.7  &68.9  &4.83  &76.0  &66.9  & 6.53 & 78.3 & 69.0 \\
\bottomrule
\end{tabular}%
}
\caption{Model Evaluation Results across Dialects.
\textbf{RL} denotes ROUGE-L (whitespace tokenization); \textbf{BF$^D_1$} denotes BERTScore-F1 (mDeBERTa); \textbf{BF$^C_1$} denotes CAMeLBERTScore-F1. }

\end{table*}

\begin{figure*}[h]
    \centering
    \includegraphics[width=0.7\linewidth]%
    {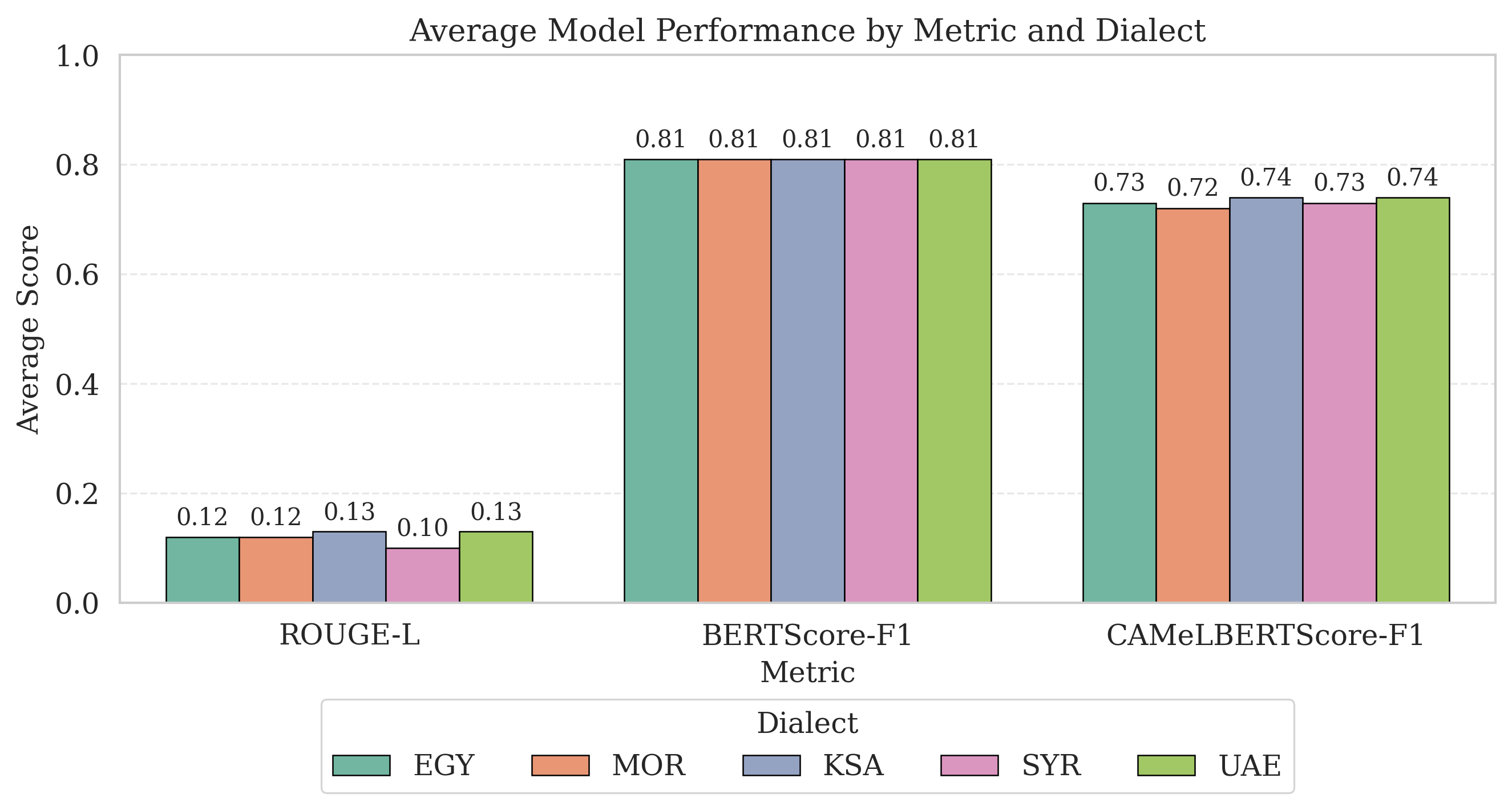}

    \caption{Performance per dialect, using the average performance for all the models.}

    \label{fig:average_perf_per_dial}
\end{figure*}

\begin{table*}[h!]
\centering
\label{tab:model_dialect_rankings}
\scalebox{0.9}{
\small
\begin{tabular}{lclcccc}
\toprule
\textbf{Model Name} & \textbf{EGY}  &\textbf{MOR}& \textbf{KSA}& \textbf{SYR} & \textbf{UAE} & \textbf{Avg. Rank $\uparrow$}\\
\midrule
GPT-5.4 & \textbf{1}  &2& \textbf{1}& 2 & \textbf{1} & 1.4 \\
Atlas-Chat-9B & 3  &\textbf{1} & 3 & \textbf{1} & 3  & 2.2\\
Gemma4-31B & 5  &4 & 2 & 6 & 2  & 3.8\\
Qwen2.5-7B & 4  &6 & 4 & 4 & 4 & 4.4 \\
Nile-Chat-4B & 2  &10 & 5 & 3 & 5 & 5.0 \\
Jais-13B-Chat & 6  &7 & 6 & 5 & 6 & 6.0\\
Qwen3.6-Plus & 8  &5& 7& 9 & 7 & 7.2\\
Command-R7B & 7  &8 & 8 & 7 & 8  & 7.6\\
Atlas-Chat-2B & 9  &3 & 9 & 8 & 9  & 7.6\\
SILMA-9B & 10  &9 & 10 & 10 & 10  & 9.8\\
Hala-9B & 11  &11 & 11 & 11 & 11  & 11.0\\
\bottomrule
\end{tabular}%
}
\caption{Model rankings across five Arabic dialects, sorted by the average rank (Lower is better).} 
\end{table*}

\end{onecolumn}

\twocolumn

\begin{onecolumn}

\section{QA Generation: Initial Prompt}
\label{app:initial_prompt}
\noindent\fbox{%
  \begin{minipage}{\dimexpr\linewidth-2\fboxsep-2\fboxrule\relax}
    \ttfamily\small
combined\_initial\_prompt = ChatPromptTemplate.from\_template("""
You are a linguist and \{passage\_language\} native speaker.
You need to generate reading comprehension non-opinionated question with precise, unambiguous answer, along with a list of exact quotes from the passage that were used to form the answer. The questions should be based on understanding and interpreting the text in the passage. A question can sometimes be derived partly through common knowledge in \{country\} that is not explicitly present in the passage.\\

It is possible to either find the answer to this question directly in the text or to infer the answer from the overall meaning of the passage. In order to answer an inferential question, pieces of information in the passage may need to be connected, or the sequence of events may need to be tracked. The answers to the generated questions must undergo all the following cases:\\

1.The answer must be found directly in the passage. It should match the passage closely in meaning but does not need to be copied verbatim. Minor adjustments are allowed for pronoun consistency (e.g., change “I did” to “he did” or “she did”), as well as slight rephrasing or morphological changes.\\

2. The question must be phrased strictly in the third person.\\

3.The answer must be inferred by piecing together information from different parts of the passage.\\

4.The answer must be inferred by interpreting the overall meaning of the passage.\\

You must not use a question similar to the previous questions. The answer to the question should not be the same as the answer to previous questions.\\

Identify and extract the full quote(s) where the answer to the question is based.\\
The quotes must be exact, and the character indices of each quote within the passage must be provided.\\
The quotes must not include any extra details that are not explicitly in the answers
Formulate each question strictly in the third person.
Questions and answers must be generated using ONLY the \{passage\_language\}. Your response must contain the question and answer only, or “N/A” if the question cannot be created.\\

Keep each question short and concise.\\

The answer must be short.\\

IMPORTANT: The question, answer and extracted sentence must be in \{passage\_language\}.Do not use MSA or other dialects, otherwise you will be penalized \$1000 per word.\\

 Return your response in the following JSON format:\\
 \{\{\\

\hspace*{.5cm}"Question": "<your question in \{passage\_language\} or N/A>",\\
\hspace*{.5cm}"Answer": "<your answer in \{passage\_language\} or N/A>",\\
\hspace*{.5cm}"Quotes":[\\
\hspace*{1cm}\{\{\\
\hspace*{1.2cm} "text": "<your quote in \{passage\_language\} or N/A>",\\
          \hspace*{1.2cm} "start\_char": <the starting character index of the quote in the passage as an integer, or \\ \hspace*{3.5cm}$-1$ if N/A>,\\
          \hspace*{1.2cm} "end\_char": <the ending character index of the quote in the passage as an integer, or\\ \hspace*{3.5cm}$-1$ if N/A>\\
\hspace*{1cm}\}\}\\
   ]\\
 \}\}\\

Passage:\\
--------------------------------\\
\{passage\}\\
--------------------------------\\

Previous Questions:\\
\{previous\_questions\}\\
""")

  \end{minipage}%
}

\end{onecolumn}

\begin{onecolumn}

\section{QA Generation: LLM-as-a-Judge Prompt}
\label{app:judge_prompt}
\noindent\fbox{%
  \begin{minipage}{\dimexpr\linewidth-2\fboxsep-2\fboxrule\relax}
    \ttfamily\small
combined\_judgement\_prompt = ChatPromptTemplate.from\_template("""\\
You are \{passage\_language\} native speaker.
You are evaluating a reading comprehension question and its answer based on the passage below, which is written in \{passage\_language\}.
Your task is to set of critical criteria and nice\-to\-have recommendations.

Label these quality and formatting dimensions (True/False):
- IsNonOpinionated: The question is objective and not based on personal opinion.
- UnambiguousAnswer: The answer is clear and does not allow multiple interpretations.
- IsUnbiased: The question is free from prejudice.
- IsAnswerable: The question can be answered from the text with the allowed outside knowledge.
- IsRelevant: The question is clearly related to the passage's content.
- IsInThirdPerson: The question is phrased strictly in the third person.
- QuestionFreeFromLinguisticOrGrammarTerms: Grammar or linguistic terms are not used in the question.
- IsShortQuestion: The question is 20 words or fewer.
- IsPreciseAnswer: The answer does not contain unnecessary details or irrelevant information.
- IsIn\{passage\_language\}: The question, answer and quotes must be in \{passage\_language\}.

Critical: If any of the checks for boolean dimensions are false, describe how to fix it. 
NiceToHave: Suggest briefly how to increase the reasoning complexity of the question.

Output your evaluation strictly as the following JSON schema. Return only the JSON object, nothing else.

Provide evaluation strictly in JSON:

\{\{\\
  "IsNonOpinionated": true|false,\\
  "IsNonOpinionated\_reason": "string",\\
  "UnambiguousAnswer": true|false,\\
  "UnambiguousAnswer\_reason": "string",\\
  "IsUnbiased": true|false,\\
  "IsUnbiased\_reason": "string",\\
  "IsAnswerable": true|false,\\
  "IsAnswerable\_reason": "string",\\
  "IsRelevant": true|false,\\
  "IsRelevant\_reason": "string",\\
  "IsInThirdPerson": true|false,\\
  "IsInThirdPerson\_reason": "string",\\
  "QuestionFreeFromLinguisticOrGrammarTerms": true|false,\\
  "QuestionFreeFromLinguisticOrGrammarTerms\_reason": "string",\\
  "IsShortQuestion": true|false,\\
  "IsShortQuestion\_reason": "string",\\
  "IsPreciseAnswer": true|false,\\
  "IsPreciseAnswer\_reason": "string",\\
  "IsIn\{passage\_language\}": true|false,\\
  "IsIn\{passage\_language\}\_reason": "string",\\
  
  "Recommendations": \{\{\\
    "Critical": "string",                // string; empty if no critical issues\\
    "NiceToHave": "string"               // string; suggestion to increase complexity\\
 \}\}\\
\}\}\\

Return only the JSON object, nothing else. Otherwise you will be penalized \$1000 per word.\\

Passage:\\
--------------------------------\\
\{passage\}\\
--------------------------------\\

Question:\\
\{question\}

Answer:
\{answer\}

Quotes:\\
\{quotes\}\\
""")

  \end{minipage}%
}

\end{onecolumn}
\begin{onecolumn}
\onecolumn
\section{QA Generation: Improvement Prompt}
\label{app:improv_prompt}

\noindent\fbox{%
  \begin{minipage}{\dimexpr\linewidth-2\fboxsep-2\fboxrule\relax}
    \ttfamily\small
combined\_improvement\_prompt = ChatPromptTemplate.from\_template(``````You are a linguist and \{passage\_language\} native speaker. Revise the original\_question and original\_answer based on the provided judge feedback.Your primary mandate is to resolve all issues in Recommendations.Critical. Systematically correct every dimension flagged as false in the feedback, using the associated \_reason fields to guide your edits. You need to generate a new version of reading comprehension non-opinionated question with precise, unambiguous answer, along with a list of exact quotes from the passage that were used to form the answer. The questions should be based on understanding and interpreting the text in the passage. A question can sometimes be derived partly through common knowledge in \{country\} that is not explicitly present in the passage. It is possible to either find the answer to this question directly in the text or to infer the answer from the overall meaning of the passage. In order to answer an inferential question, pieces of information in the passage may need to be connected, or the sequence of events may need to be tracked. The answers to the generated questions must undergo all the following cases:\\

1.The answer must be found directly in the passage. It should match the passage closely in meaning but does not need to be copied verbatim. Minor adjustments are allowed for pronoun consistency (e.g., change “I did” to “he did” or “she did”), as well as slight rephrasing or morphological changes.
2. The question must be phrased strictly in the third person.
3.The answer must be inferred by piecing together information from different parts of the passage.
4.The answer must be inferred by interpreting the overall meaning of the passage.

Identify and extract the full quote(s) where the answer to the question is based.
The quotes must be exact, and the character indices of each quote within the passage must be provided.
The quotes must not include any extra details that are not explicitly in the answers Formulate each question strictly in the third person.
Questions and answers must be generated using ONLY the \{passage\_language\}.
Your response must contain the question and answer only, or ``N/A'' if the question cannot be created.
Keep each question short and concise.
The answer must be short.

IMPORTANT: The question, answer and extracted sentence must be in \{passage\_language\}.Do not use MSA or other dialects, otherwise you will be penalized \$1000 per word.

Return your response in the following JSON format:
 \{\{
\\  
\hspace*{1cm}\\"Question\\": \\"<your question in \{passage\_language\} or N/A>\\",\\
\hspace*{1cm}"Answer": \\"<your answer in \{passage\_language\} or N/A>\",\\
\hspace*{1cm}\\"Quotes\\":[
\\  \hspace*{2cm}      \{\{\\
\hspace*{2.5cm}\\"text\\": "<your quote in \{passage\_language\} or N/A>\",\\\hspace*{2.5cm}"start\_char": <the starting character index of the quote in the passage as an integer,\\
\hspace*{4.5cm} or $-1$ if N/A>,\\
\hspace*{2.5cm}"end\_char": <the ending character index of the quote in the passage as an integer, \\\hspace*{4.5cm}or $-1$ if N/A>\\
 \hspace*{2cm}       \}\}\\
  ]\\
  \}\}\\

Passage:\\
--------------------------------\\
\{passage\}\\
--------------------------------\\

Original Question: \\
\{original\_question\}\\

Original Answer: \\
\{original\_answer\}\\

Original Quotes:\\
\{original\_quotes\}\\

Judge Feedback:\\
\{judge\_feedback\}\\
'''''')

  \end{minipage}%
}

\end{onecolumn}

\twocolumn
\section{Text Structure Restoration from Dialectal Arabic Transcripts}
\label{app:text-restoration}

\begin{strip}
    \begin{center}
    \includegraphics[width=.96\textwidth,
    height=.32\textheight,
    keepaspectratio]{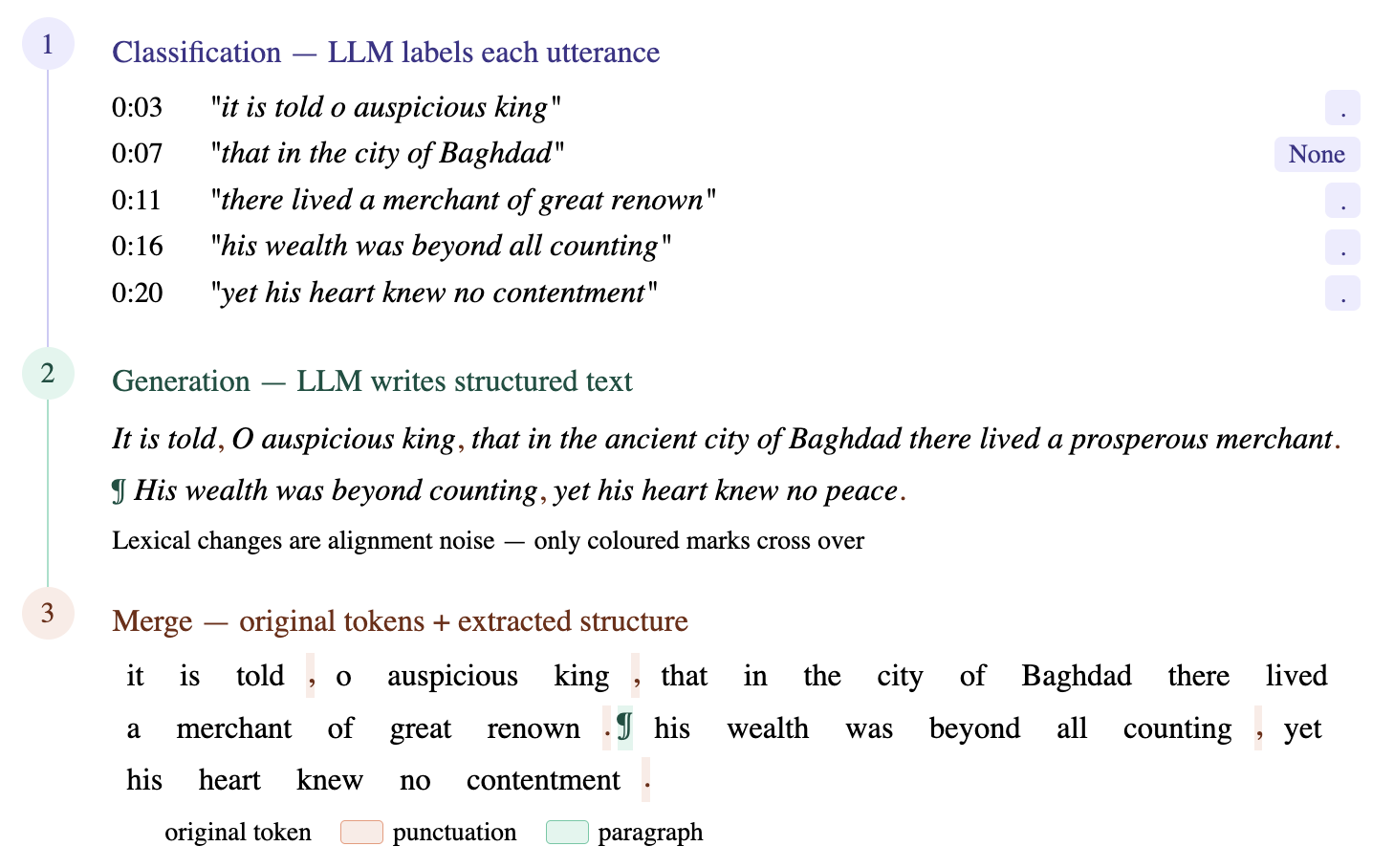}
    \captionof{figure}{Three-stage pipeline for restoring punctuation and paragraph structure from raw transcripts.}  
    \label{fig:text-restoration}
    \end{center}
    
\end{strip}

Given raw transcripts with utterances and timings, we restore this structure (sentence split, punctuation, paragraphs) in three stages, as shown in Figure~\ref{fig:text-restoration}.

\paragraph{Stage~1: Utterance-level classification}
The input transcript is represented as utterances (sequences of tokens) with speech timings.
The LLM maps each utterance to a punctuation label or \texttt{None}.
The list of utterances is presented to an LLM together with their time intervals, and the model assigns a punctuation label or  \texttt{None}. As recent research shows, such a representation of the task is more natural for generation models than the token-based alternative. Results of classification allow establishing raw text boundaries based on non-verbal signs (pauses) and contextual information for the next steps.
 
\paragraph{Stage~2: Generation of structured text}
The utterances, along with labels, are concatenated and passed to the LLM, which generates coherent, punctuated, paragraph-structured text.
This step allows the model to correct ambiguities -- for example, a pause-final utterance initially marked as a full stop may become a comma when the following utterance begins a continuation clause, or a new paragraph may be inserted  where a shift of topic is detected.
In addition, the model adds punctuation within an utterance where two clauses run together in speech without a pause.

However, the generated text may differ lexically from the original (added words, dropped particles, paraphrases). Such differences are text noise that is cleaned in the next stage.

\paragraph{Stage~3: Merge with original tokens}
A sequence matching algorithm \cite{ratcliff1988pattern} is performed between the generated tokens and the original tokens.
The alignment walks through the tokens in order, identifying the position of each punctuation mark and paragraph break relative to the nearest preceding original token. Then, only punctuation and paragraph markers are extracted and inserted at the corresponding positions in the original token stream.

\section{Additional Details on the Experimental Setup}
\label{app:experimental}
A list of the evaluated language models and their sizes is provided in Table~\ref{tab:List_models}. The sizes of GPT-5.4 and Qwen3.6-Plus models are not publicly disclosed. For most models, we used OpenRouter to perform inference, resulting in a total cost of approximately \$200. However, not all models were available on this platform, particularly the Arabic-centric ones. For those, we ran inference on a single NVIDIA RTX 6000.
\begin{onecolumn}
\begin{table}[h]
\footnotesize
    \centering
    \begin{tabular}{llllrcc}
\toprule
 & \textbf{Family} & \textbf{Model ID} & \textbf{Short Name}&\textbf{Size (B)} & \textbf{Ar} & \textbf{En} \\
\midrule
1  & MBZUAI-Paris & Atlas-Chat-2B & Atlas-Chat-2B& 2 & $\bullet$ &   \\
2  & MBZUAI-Paris & Atlas-Chat-9B & Atlas-Chat-9B & 9 & $\bullet$ &   \\
3  & hammh0a & Hala-9B & Hala-9B & 9 & $\bullet$ &   \\
4  & MBZUAI-Paris & Nile-Chat-4B & Nile-Chat-4B & 4 & $\bullet$ &   \\
5  & inceptionai & Jais-13B-Chat & Jais-13B-Chat & 13 & $\bullet$ & $\bullet$ \\
6  & Qwen & qwen3.6-plus & Qwen3.6-Plus & N/A & $\bullet$ & $\bullet$ \\
7  & Qwen & Qwen2.5-7B-Instruct & Qwen2.5-7B & 7 & $\bullet$ & $\bullet$ \\
8 & CohereLabs & c4ai-command-r7b-arabic-02-2025 & Command-R7B & 7 & $\bullet$ & $\bullet$ \\
9 & silma-ai & SILMA-9B-Instruct-v1.0 & SILMA-9B & 9 & $\bullet$ & $\bullet$ \\
10  & google & gemma-4-31B-it & Gemma4-31B &  31 & $\bullet$ & $\bullet$ \\
11  & openai & GPT-5.4 & GPT-5.4 & N/A & $\bullet$ & $\bullet$ \\

\bottomrule
\end{tabular}
    \caption{A list of the evaluated language models and their sizes.}\label{tab:List_models}
\end{table}
\end{onecolumn}

\section{Final dataset stats}
\label{app:final_stats}

\begin{table}[htbp]
\centering
\small
\begin{tabular}{lrrrrrrr}
\toprule
\textbf{Dialect} & \textbf{Initial Pairs}& \textbf{Flagged} & \textbf{Fix} & \textbf{Discard} & \textbf{No Change} & \textbf{Final Pairs} \\
\midrule
EGY & 1000 & 41 (4.1\%) & 17 & 1 & 23 & 999 \\
MOR & 1000 & 31 (3.1\%) & 20 & 3  & 8 & 997\\
KSA    & 1000 & 55 (5.5\%) & 27 & 1 & 27 & 999\\
SYR   & 1000 & 60 (6.0\%) & 33 & 16  & 11 & 984\\
UAE  & 1000 & 22 (2.2\%) & 5  & 2  & 15 & 998\\
Total & 5000 & 209 (4.2\%) & 102 & 23 & 84 & 4977\\
\bottomrule
\end{tabular}
\caption{Post-review reconciliation of the total number of QA pairs before and after review per dialect.}
\label{tab:dialect_fixes_total}
\end{table}

\section{Genre Descriptions} 

\label{sec:appendix-genre-descriptions}
\begin{table}[!ht]
\centering

\small

\renewcommand{\arraystretch}{1.2}
\begin{tabularx}{\linewidth}{>{\raggedright\arraybackslash}p{4.2cm} >{\raggedright\arraybackslash}X}
\toprule
\textbf{Genre} & \textbf{Description} \\
\midrule
\textbf{Personal Development} & Focuses on everyday living, well-being, hobbies, and individual growth. \\[4pt]
\textbf{Sports} & Athletic activities, organized games, and competitive event analysis. \\[4pt]
\textbf{History} & Dedicated to exploring and preserving historical events and cultural backgrounds. \\[4pt]
\textbf{Interviews} & First-person accounts, direct human interaction, and conversational dialogue. \\[4pt]
\textbf{Documentary} & Non-fiction, real-world observation, and deep-dive reporting. \\[4pt]
\textbf{Arts, Culture \& Entertainment} & Creative performances, cultural trends, and media critique. \\[4pt]
\textbf{Science, Tech \& Innovation} & Explores the natural world, scientific inquiry, and technological advancements. \\[4pt]
\textbf{Education} & Educational texts, instructions, and publicly available information sharing. \\[4pt]
\textbf{Business \& Economy} & Commerce, financial markets, macroeconomic theories, and business stories. \\[4pt]
\textbf{Other} & News (Politics, Society), fiction (Drama, Comedy, Romance, Youth) etc. \\
\bottomrule
\end{tabularx}
\caption{YouTube video genres in EDRAC.}
\label{tab:genre_descriptions}
\end{table}
\begin{figure*}[t]
    \centering

    \begin{subfigure}[t]{0.45\textwidth}
        \centering
        \includegraphics[width=\linewidth]{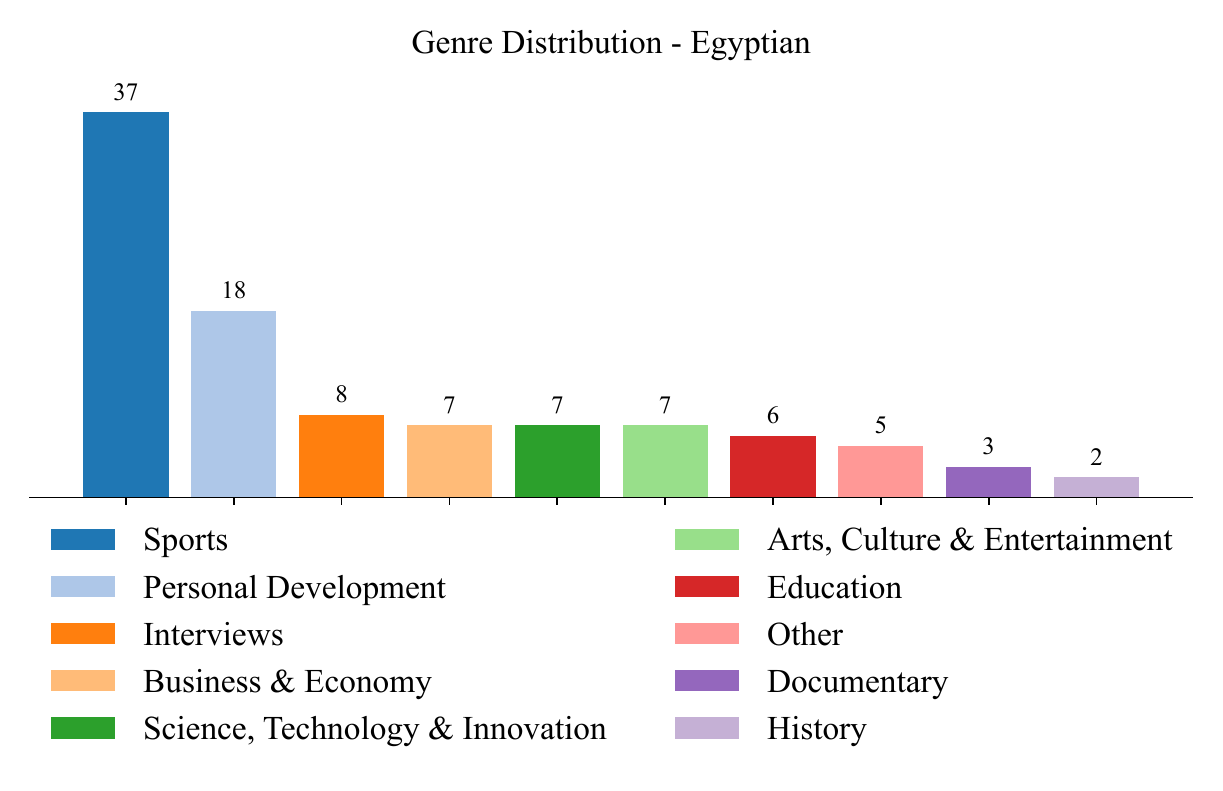}
        \label{fig:genre-egy}
    \end{subfigure}
    \hfill
    \begin{subfigure}[t]{0.45\textwidth}
        \centering
        \includegraphics[width=\linewidth]{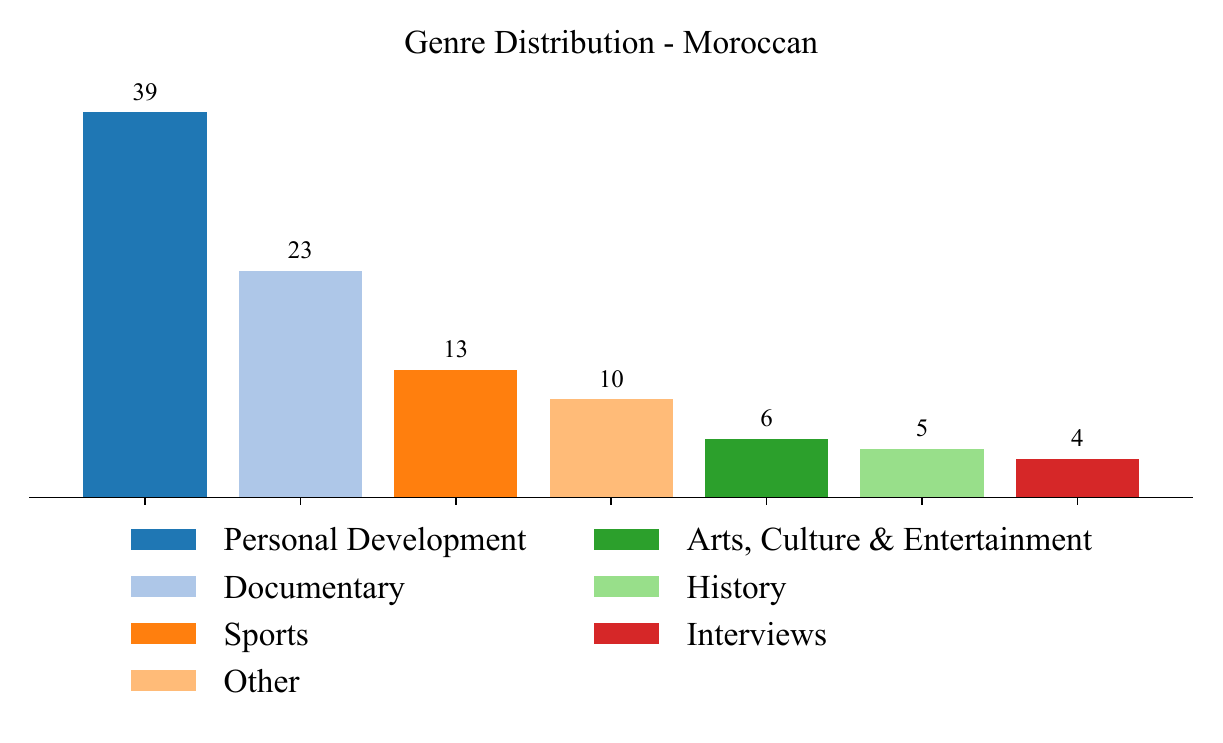}
        \label{fig:genre-mor}
    \end{subfigure}

    \vspace{2mm}

    \begin{subfigure}[t]{0.45\textwidth}
        \centering
        \includegraphics[width=\linewidth]{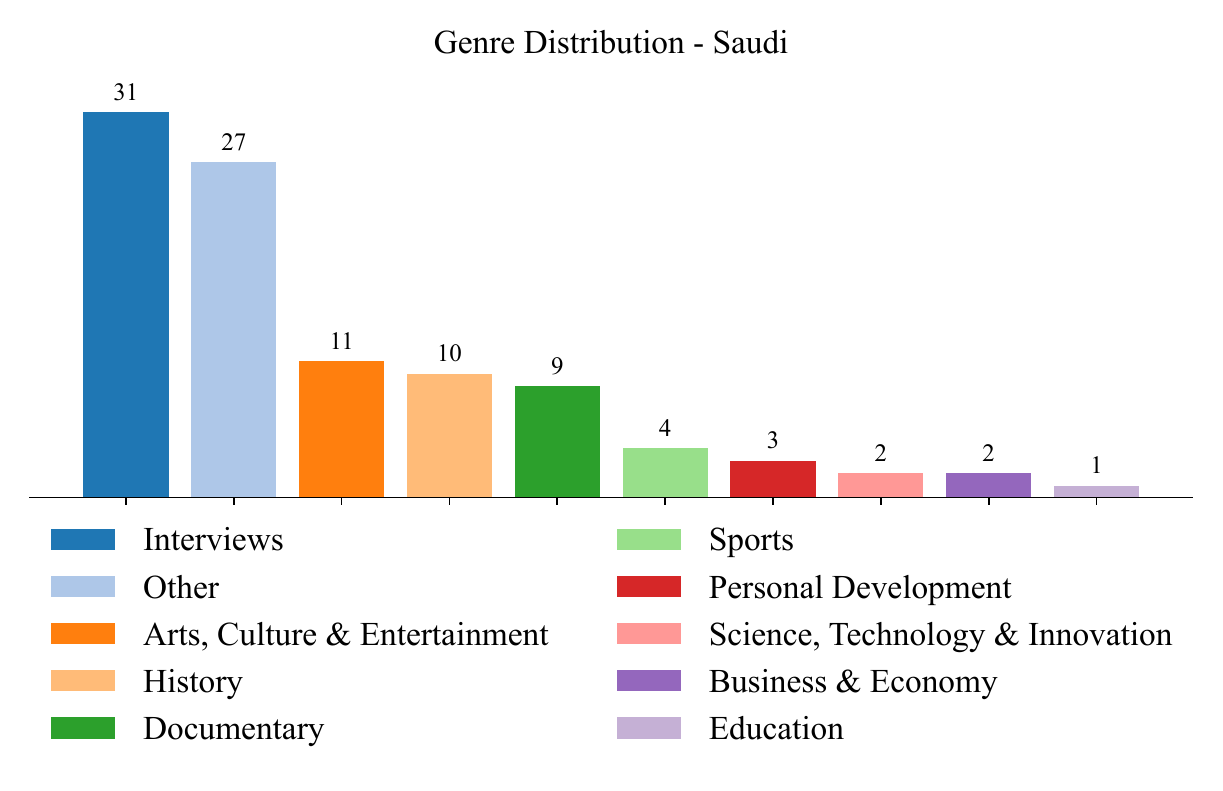}
        \label{fig:genre-sau}
    \end{subfigure}
    \hfill
    \begin{subfigure}[t]{0.35\textwidth}
        \centering
        \includegraphics[width=\linewidth]{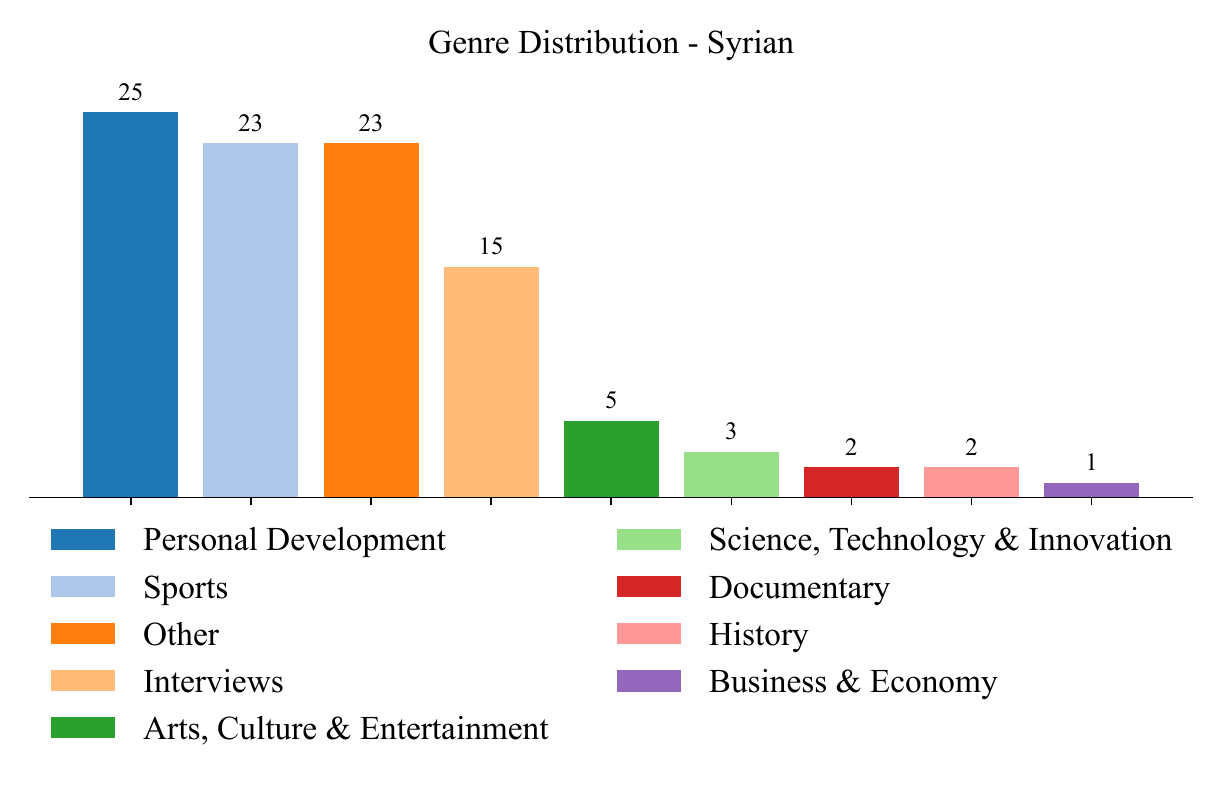}
        \label{fig:genre-syr}
    \end{subfigure}

    \vspace{2mm}

    \begin{subfigure}[t]{0.45\textwidth}
        \centering
        \includegraphics[width=\linewidth]{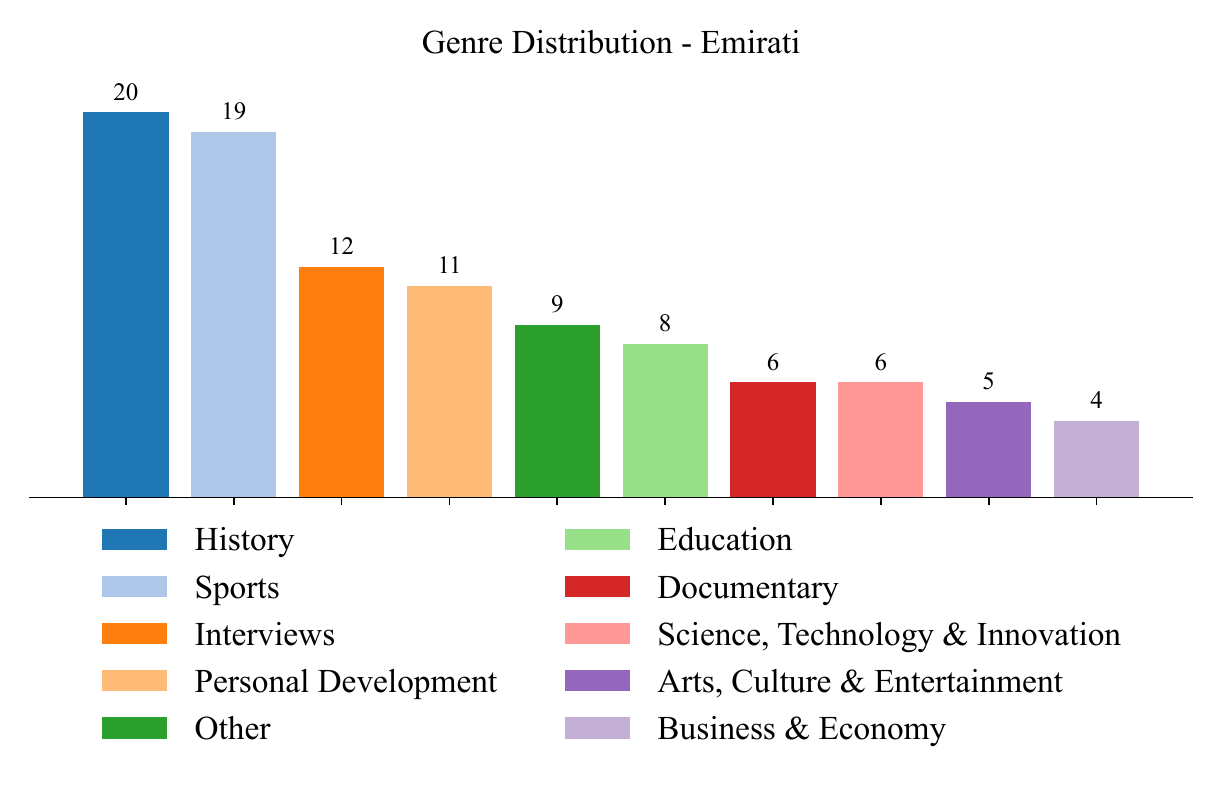}
        \label{fig:genre-ema}
    \end{subfigure}

    \caption{Genre distribution by dialect: Egyptian, Moroccan, Saudi, Syrian and Emirati.}
    \label{fig:genre-distribution-by-dialect}
\end{figure*}

\clearpage
\section{Channels per dialect} \label{app:channel}

\begin{table}[htbp]
\centering
\label{tab:unique_channels_arabtex}
\small
\setlength{\tabcolsep}{4pt}
\renewcommand{\arraystretch}{1.15}
\begin{tabularx}{\linewidth}{
    @{}l >{\raggedright\arraybackslash}X@{}
}
\toprule
\textbf{Dialect} & \textbf{List of Channels} \\
\midrule
EGY & Ahmed Waheed, Al Nahar Drama, El-Podcasters - \RL{البودكاسترز}, Film Gamed, Kareem Esmail, MahmoudIsmailTV, Mamdouh NasrAllah, Omar Khaled - \RL{عمر خالد},\\
& Pharmastan - \RL{فارماستان}, ReadTube - \RL{جيل يقرأ}, STUDIO 77, Sarah Abouelkhair,Sarah Hany, Sherif Nabil - \RL{بتاع أكشن}, Waleed - \RL{وليد}, joo sport \RL{الكورة في السريع}, \\
&\RL{أسئلة نص الليل}, \RL{المخبر الاقتصادي} - Mokhbir Eqtisadi, \RL{تاريخ و كورافيا - شادى حبشى}, \RL{دروس أونلاين}, \RL{عشوائيات}, \RL{فاهم بودكاست}, \RL{هبة ابو الخير} Heba Abo Elkheir \\
\midrule
MOR & BarbaRoss Hicham \RL{برباروس هشام}, Farouk Life , Foot Maroc \textbar\ \RL{فووت ماروك}, Hassan El Fad \textbar\ \RL{حسن الفد}, IBra Traveler, OTMAN HANA / \RL{عثمان حانا}, Said Naciri \\
&\RL{سعيد الناصري}, Yassin Haro, ahssan patissier, elbachir L3aouni \RL{البشير العوني}, soumamalak, \RL{إبن فطومة} Ibnfatoma , \RL{نكهة مغربية} \\
\midrule
KSA & Alaab - \RL{ألعاب}, SBC Channel, \RL{السعودية}, \RL{هاي فايف} \textbar\ Highfive \\
\midrule
SYR & \RL{بانة السورية}, Al Mashhad Light \RL{المشهد لايت}, Barhom m3arawi - \RL{برهوم معراوي}, Boshe Tv, CHEF OMAR \RL{شيف عمر}, Donia Stories - \RL{اسمعولي هالقصة}, Ghaith Marwan \\
&\RL{غيث مروان}, LTV \RL{تلفزيون}, Majed Haidar - \RL{ماجد حيدر}, NewDose - \RL{نيودوس}, Rano's Home , Sama Art International \textbar\ \RL{سامه للإنتاج الفني}, Sara alwari \RL{سارة الورع}, Sherin Amara, \\
&Syria TV , Syrian kitchen, Yala Story Plus, \RL{الثانية} - Althania, \RL{الريف السوري فلوجر}, \RL{الشيف الشامي} - ShamiChef , \RL{جلال و ياسمين} \textbar\ Jalal \& Yasmin , \RL{جهينة هوم} Johina Home,\\
&\RL{حواديت السيما}, \RL{سوري جيمر} - Syrian Gamer, \RL{قناة حلب اليوم} Halab Today TV \\
\midrule
UAE & 4042 Studios, Abu Dhabi TV \RL{قناة أبوظبي}, Hamdan Al-Ali, UAE FALCONS FEDERATION, \RL{تلفزيون دبي} - Dubai TV, \RL{خالد الخالدي} Khaled Alkhaaldi, \RL{رنا} Ranoy7, \RL{سما دبي}\\
&- Sama Dubai, \RL{قناة الظفرة}, \RL{مركز حمدان بن محمد لإحياء التراث} \\
\bottomrule
\end{tabularx}
\caption{Final list of YouTube Channels by Dialect.}

\end{table}

\begin{onecolumn}

\includepdf[pages=1, frame, scale=0.75,pagecommand=\section{Transcription Guidelines}\label{app:transc_guide}]{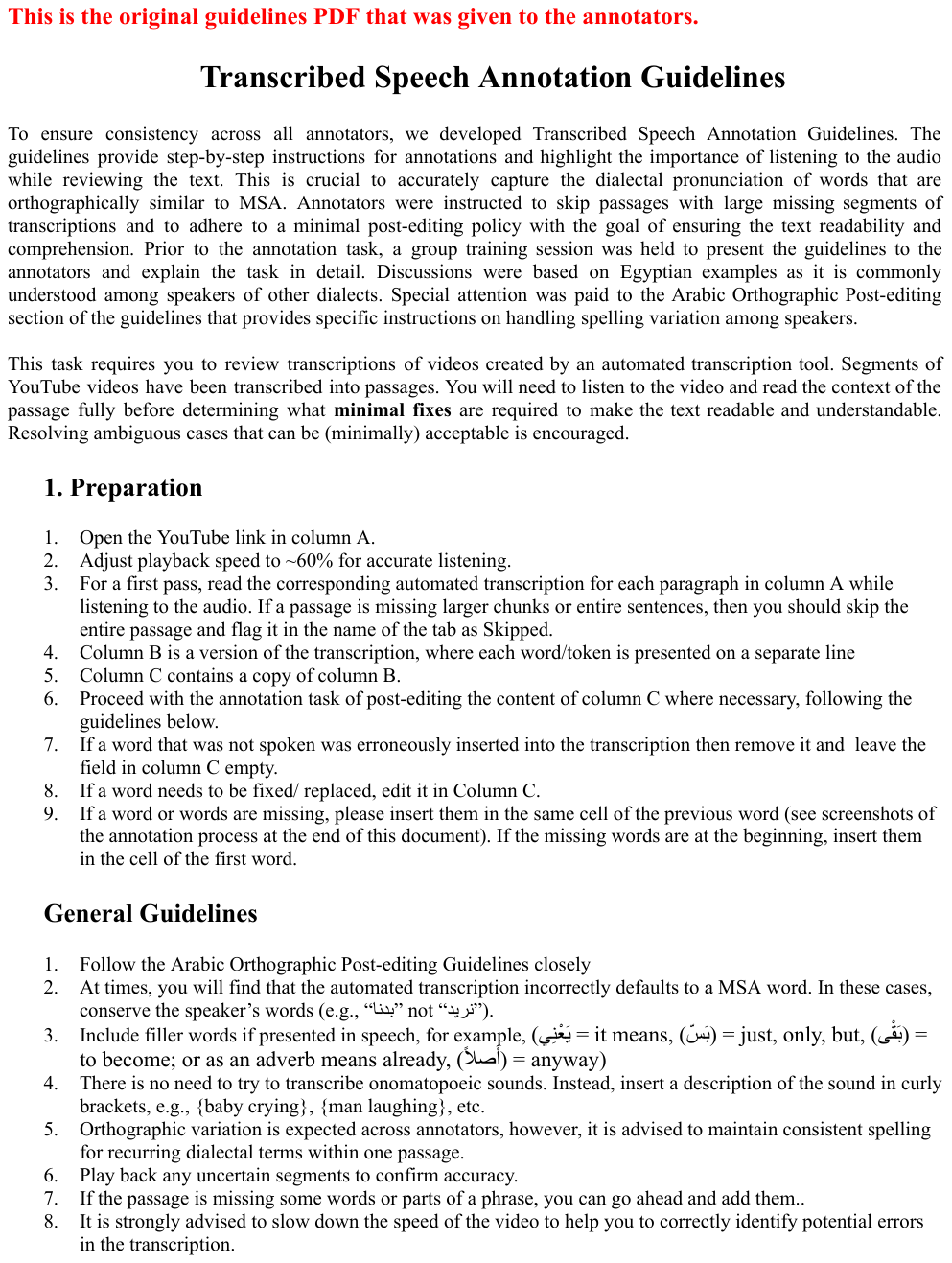}
\includepdf[pages=2-, scale=0.75,frame]{Transcribed_Speech_Annotation_Guideline_n.pdf}

\end{onecolumn}
\begin{onecolumn}
\includepdf[pages=1, frame, scale=0.75,pagecommand=\section{Annotation Guidelines}\label{app:qa_guide}]{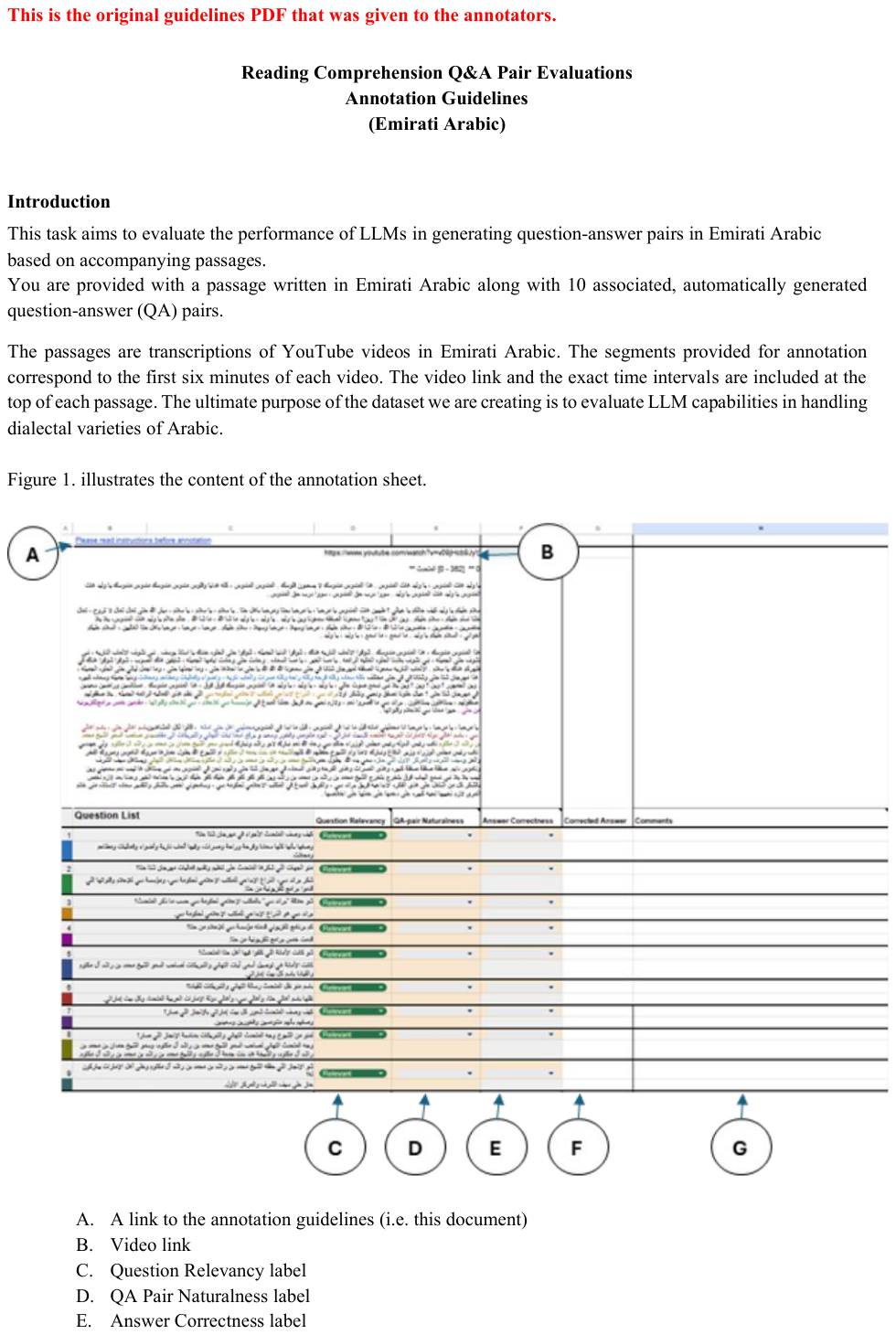}
\includepdf[pages=2-, frame, scale=0.75]{Emirati_Guidelines_EN_fixed.pdf}

\end{onecolumn}

\section{Annotators' Demographics}\label{app:demographics}
\begin{table*}[!ht]
\centering
\small
\begin{tabular}{lllllllll}
\toprule
\textbf{ID} & \textbf{Dataset} & \textbf{Native} & \textbf{Residence} & \textbf{Age} & \textbf{Gender} & \textbf{Degree} & \textbf{Task} & \textbf{Background} \\
\midrule
A1 & EGY & EGY & EGY & 30s & F & BA  & A, R & LA, PT \\
A2 & EGY & SYR & SYR & 40s & F & PhD & A, R & CT \\
A3 & EGY & EGY & EGY & 30s & F & BA  & A    & LA, RA, CT \\
A4 & EGY & EGY & EGY & 30s & F & PhD  & A,R & L, CL \\
A5 & EGY & EGY & EGY & 40s & F & MA & A,R & L, AL \\
\midrule
A6 & MOR &MOR & MOR & 30s & F & BA & A     & RA, LA \\
A7 & MOR & MOR & MOR & 30s & F & BA & A, R  & LA \\
A8 & MOR & MOR & MOR & 40s & M & PhD & A   & LA, RA, CT \\
A9 & MOR & MOR & MOR & 30s & M & MBA & A, R & CT, PT \\
A10 & MOR & MOR & FR & 30s & F & PhD & A,R & CS \\
A11 & MOR & MOR & MOR & 30s & F & BA & A,R  & EL \\
\midrule
A12 & KSA & SYR & SYR & 40s & M & BA& A, R & LA, RA, CT \\
A13 & KSA & SYR & SYR & 40s & F & BA & A, R & LA \\
A14 & KSA & SYR & SYR & 50s & F & BA & A    & LA, PT \\
A15 & KSA & KSA & KSA & 30s & F & PhD& A,R & TR \\
A16 & KSA & KSA & KSA & 30s & F & PhD & A,R & CL \\
\midrule
A17 & SYR & SYR & SYR & 40s & M & BA & A    & LA, RA, CT \\
A18 & SYR & SYR & SYR & 20s & F & BA & A, R & CT, RA \\
A19 & SYR & SYR & SYR & 30s & F & BA & A, R & CT, PT \\
A20 & SYR & SYR & SYR & 40s & F & BA & A    & LA, RA, PT \\
A21 & SYR & SYR & UK & 40s & M & MA & A,R   & LA \\
A22 & SYR & SYR & SYR & 20s & M & MS & A,R & CS \\
\midrule
A23 & UAE & PAL & UAE & 20s & M & BA & A, R & LA \\
A24 & UAE & PAL & UAE & 40s & M & BA & A & LA \\
A25 & UAE & PAL & UAE & 40s & F & BA & A, R & LA \\
A26 & UAE & UAE & USA & 40s & F & BA & A,R & LA \\
A27 & UAE & UAE & UAE & 20s & M & BA & A,R& CS \\
\bottomrule
\end{tabular}
\caption{Annotators' demographics and roles for the transcription correction task. There are two roles: Annotate (A), and Review (R). The annotators' background experience includes: Certified Teacher (CT), Private Tutor (PT), Linguistic Annotator (LA), and Research Assistant (RA). All annotators
are native speakers of Arabic, and worked on their native dialect or dialects of a country
where they have resided for more than 15 years (if they worked on a different dialect).}

\label{tab:annotators}

\end{table*}

\begin{table*}[!ht]
\centering
\small
\begin{tabular}{lllllllll}
\toprule
\textbf{ID} & \textbf{Dataset} & \textbf{Native} & \textbf{Residence} & \textbf{Age} & \textbf{Gender} & \textbf{Degree} & \textbf{Task} & \textbf{Background} \\
\midrule
A1 & EGY & EGY & EGY & 30s & F & PhD  & A,R & L, CL \\
A2 & EGY & EGY & EGY & 40s & F & MA & A,R & L, AL \\
\midrule
A3 & MOR & MOR & FR & 30s & F & PhD & A,R & CS \\
A4 & MOR & MOR & MOR & 30s & F & BA & A,R  & EL \\
\midrule
A5 & KSA & KSA & KSA & 30s & F & PhD & A,R & CL \\
A6 & KSA & KSA & KSA & 20s & F & BA & A,R & CS \\
\midrule
A7 & SYR & SYR & UK & 40s & M & MA & A,R   & LA \\
A8 & SYR & SYR & SYR & 20s & M & MS & A,R & CS \\
\midrule
A9 & UAE & UAE & USA & 40s & F & BA & A,R & LA \\
A10 & UAE & UAE & UAE & 20s & M & BA & A,R& CS \\
\bottomrule
\end{tabular}
\caption{Annotators' demographics and roles for the QA annotation task. There are two roles: Annotate (A), and Review (R). The annotators' educational background includes: Linguistics (L), Computational Linguistics (CL), Applied Linguistics (AL), English Linguistics (EL), Translation (TR) and Computer Science (CS). All annotators
are native speakers of Arabic, and worked on their native dialect or dialects of a country
where they have resided for more than 15 years (if they worked on a different dialect).}
\end{table*}

\begin{table*}[!ht]
   \centering
    \begin{tabular}{l|l}
      Task   & Payment in USD \\
      \hline
       Transcription Correction  &  \$3,540.0\\
       QA Annotation & \$4,330.0\\
    \end{tabular}
    
    \caption{Annotators were paid \$15 an hour for each one of the tasks. In the table above, we provide the total amount of money in USD paid to the annotators per task.}
    \label{tab:placeholder}
\end{table*}

\newpage

\section{IAA For QA} 

\label{sec:IAA_QA}
\begin{table}[h]
\scalebox{0.8}{
\begin{tabular}{llcccccc}
\toprule
 &  & \textbf{Cohen Kappa} &  &  & \textbf{Simple Agreement} &  &  \\
 &  & \textbf{P$_{o}$} & \textbf{P$_{e}$} & \textbf{Kappa(x)} & \textbf{Majority Disagreement} & \textbf{Majority Agreement} &  \textbf{PABAK-OS} \\
 \midrule
EGY & Relevancy & 1.00 & 1.00 & undefined & 0.00 & 1.00 & 1.00 \\
 & Naturalness & 0.93 & 0.93 & 0.00 & 0.08 & 0.93 &  0.93\\
 & Correctness & 0.97 & 0.97 & 0.00 & 0.03 & 0.97 &  0.97\\
MOR & Relevancy & 1.00 & 1.00 & undefined & 0.00 & 1.00 &  1.00\\
 & Naturalness & 0.97 & 0.97 & 0.00 & 0.03 & 0.97 &  0.97\\
 & Correctness & 0.98 & 0.98 & 0.00 & 0.01 & 0.98 &  0.99\\
KSA & Relevancy & 1.00 & 1.00 & undefined & 0.00 & 1.00 &  1.00\\
 & Naturalness & 0.92 & 0.90 & 0.15 & 0.08 & 0.91 & 0.94 \\
 & Correctness & 0.99 & 0.99 & 0.00 & 0.02 & 0.98 & 0.99\\
SYR & Relevancy & 1.00 & 1.00 & undefined & 0.00 & 1.00 &  1.00\\
 & Naturalness & 0.99 & 0.99 & 0.00 & 0.01 & 0.99 &  0.99\\
 & Correctness & 0.80 & 0.73 & 0.26 & 0.19 & 0.80 &  0.97\\
UAE & Relevancy & 1.00 & 1.00 & undefined & 0.00 & 1.00 &  1.00\\
 & Naturalness & 0.81 & 0.82 & -0.06 & 0.17 & 0.82 &  0.65\\
 & Correctness & 0.98 & 0.97 & 0.33 & 0.01 & 0.98 &  0.99\\
 \bottomrule
\end{tabular}}
\caption{Inter-Annotator Agreement for QA Annotations. K is `undefined' when P$_{o}$ = P$_{e}$ =1.}
\label{tab:IAA_agreement}
\end{table}

\section{Example of a flagged issue}
\label{app:flagged_issue}

\begin{figure*}[h!]
    \centering
    \makebox[\textwidth][c]{
        \includegraphics[trim={0.2cm 17cm 0.5
        0.2cm 0.2cm}, clip, width=1\textwidth]{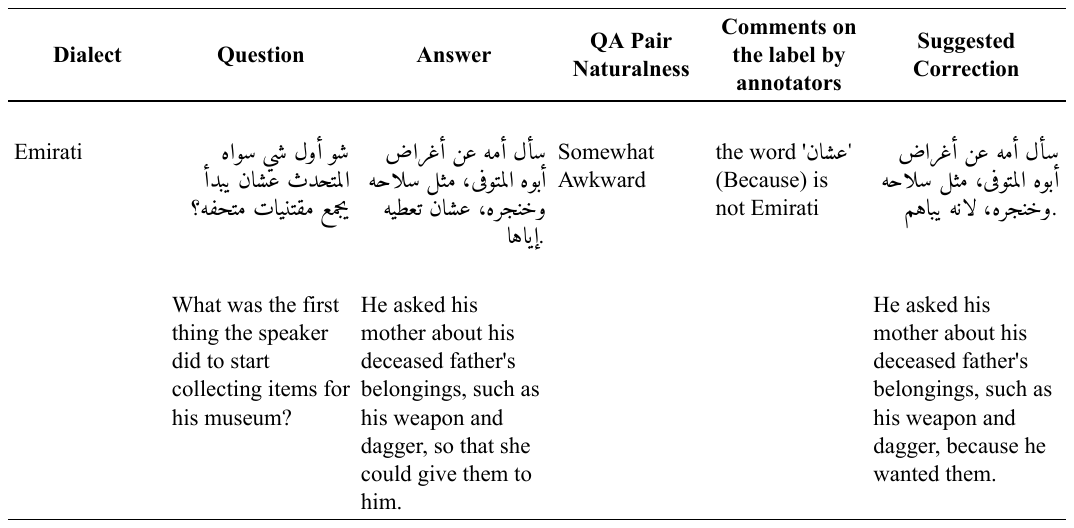}
    }
    \caption{Example of a flagged issue in Emirati Arabic by our annotators.}
    \label{fig:err_flagged}
\end{figure*}

\begin{onecolumn}

\includepdf[pages=1, frame, scale=0.75,pagecommand=\section{Human Evaluation Guidelines}\label{app:hum_eval_guide}]{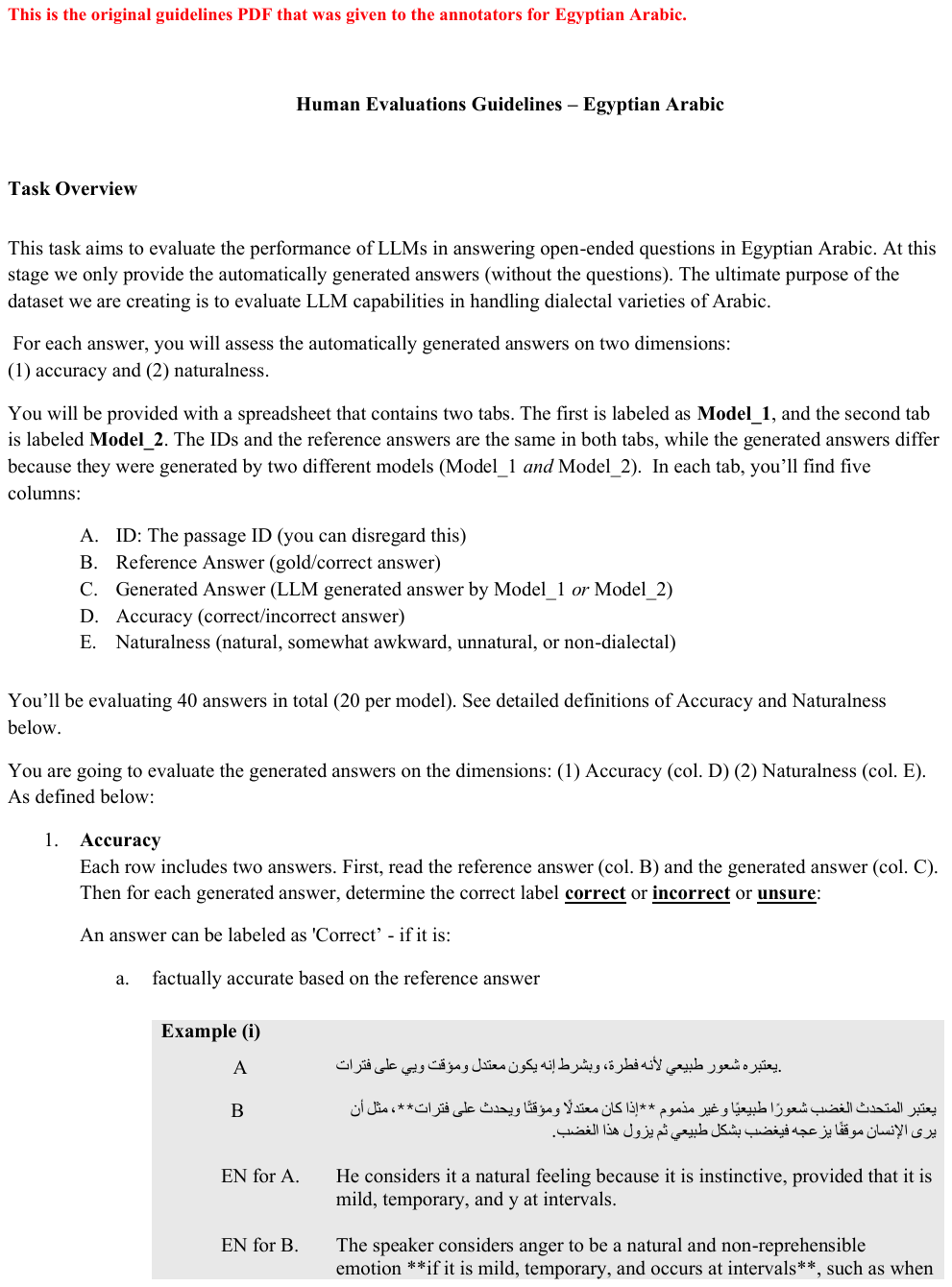}
\includepdf[pages=2-, frame, scale=0.75]{Human_Evaluations_GuidelinesEgy.pdf}

\end{onecolumn}

\end{document}